\documentclass[11pt]{article}
\usepackage{xspace}
\usepackage{ifthen}
\usepackage{fullpage}
\usepackage{times}
\usepackage[section]{definitions}
\usepackage{amsmath}
\usepackage{graphicx}
\usepackage{epsf}
\usepackage[dvips]{color}
\usepackage{algorithm}
\usepackage[noend]{algorithmic}
\usepackage{hyperref}

\newcommand{\NP}{\ensuremath{\mathrm{\mathbf{NP}}}\xspace}
\DeclareMathOperator*{\argmax}{argmax}

\newcommand{\SDSMA}{\ensuremath{\mathrm{SDS}_{\mathrm{MA}}}\xspace}
\newcommand{\SDSOMP}{\ensuremath{\mathrm{SDS}_{\mathrm{OMP}}}\xspace}

\newcommand{\LCSET}[1][]{\ensuremath{%
\ifthenelse{\equal{#1}{}}{L}{L_{#1}}}\xspace}

\newcommand{\QSET}[1][]{\ensuremath{%
\ifthenelse{\equal{#1}{}}{Q}{Q_{#1}}}\xspace}
\newcommand{\OPT}{\ensuremath{\text{OPT}}\xspace}
\newcommand{\OPTSET}[1]{\ensuremath{S^*_{#1}}\xspace}
\newcommand{\GrSet}[1]{{\ensuremath{S^G_{#1}}}\xspace}
\newcommand{\FrSet}{{\ensuremath{S^{\text{FR}}}}\xspace}
\newcommand{\OmpSet}{{\ensuremath{S^{\text{OMP}}}}\xspace}
\newcommand{\OblSet}{{\ensuremath{S^{\text{OBL}}}}\xspace}

\newcommand{\DMa}{{\ensuremath{D^{\text{MA}}}}\xspace}
\newcommand{\DOmp}{{\ensuremath{D^{\text{OMP}}}}\xspace}
\newcommand{\DOpt}{{\ensuremath{D^*}}\xspace}

\newcommand{\FD}[1]{\ensuremath{F(#1)}\xspace}
\newcommand{\FMA}[1]{\ensuremath{\hat{F}(#1)}\xspace}
\newcommand{\fma}[2]{\ensuremath{f({#1, #2})}\xspace}
\newcommand{\FMAf}{\ensuremath{\hat{F}}\xspace}

\newcommand{\FOMP}[1]{\ensuremath{\tilde{F}(#1)}\xspace}

\newcommand{\singmin}[1]{\ensuremath{\sigma_{\min}(#1)}\xspace}
\newcommand{\eigmin}[1]{\ensuremath{\lambda_{\min}(#1)}\xspace}
\newcommand{\eigmax}[1]{\ensuremath{\lambda_{\max}(#1)}\xspace}
\newcommand{\eigmink}[2]{\ensuremath{\lambda_{\min}(#1,#2)}\xspace}
\newcommand{\eigmaxk}[2]{\ensuremath{\lambda_{\max}(#1,#2)}\xspace}
\newcommand{\Superm}{\ensuremath{\gamma}\xspace}

\newcommand{\superm}[3][]{\ensuremath{%
\ifthenelse{\equal{#1}{}}{\Superm_{#2,#3}}{\Superm_{#2,#3}(#1)}}
\xspace}

\newcommand{\Eig}{\ensuremath{\lambda}\xspace}
\newcommand{\COVMS}[1]{\ensuremath{C^#1}\xspace}

\newcommand{\Corr}[2]{\ensuremath{\rho(#1, #2)}\xspace}
\newcommand{\Residual}[2]{\ensuremath{{\rm Res}(#1, #2)}\xspace}

\newcommand{\Rt}[2]{\ensuremath{R^2_{#1, #2}}\xspace}

\newcommand{\Vector}[1]{\ensuremath{{\mathbf{#1}}}\xspace}

\newcommand{\COVM}[1][]{\ensuremath{%
\ifthenelse{\equal{#1}{}}{C}{C_{#1}}}\xspace}
\newcommand{\CovM}[2]{\ensuremath{c_{#1,#2}}\xspace}
\newcommand{\COVV}[1][]{\ensuremath{%
\ifthenelse{\equal{#1}{}}{\Vector{b}}{\Vector{b}_{#1}}}\xspace}
\newcommand{\COVVS}[1]{\ensuremath{\Vector{b}^{#1}}\xspace}
\newcommand{\CovV}[1]{\ensuremath{b_{#1}}\xspace}
\newcommand{\COVMP}[1][]{\ensuremath{%
\ifthenelse{\equal{#1}{}}{{C'}}{{C'_{#1}}}}\xspace}
\newcommand{\CovMP}[2]{\ensuremath{c'_{#1,#2}}\xspace}
\newcommand{\COVVP}[1][]{\ensuremath{%
\ifthenelse{\equal{#1}{}}{\Vector{b'}}{\Vector{b'}_{#1}}}\xspace}

\newcommand{\COVMPP}[1][]{\ensuremath{%
\ifthenelse{\equal{#1}{}}{C''}{C''_{#1}}}\xspace}

\newcommand{\Coeff}[1]{\ensuremath{\alpha_{#1}}\xspace}

\newcommand{\VM}{\ensuremath{A}\xspace}
\newcommand{\VMC}{\ensuremath{A_{\rho}}\xspace}

\newcommand{\COVMC}{\ensuremath{C_{\rho}}\xspace}

\newcommand{\COEFFVEC}[1][]{\ensuremath{%
\ifthenelse{\equal{#1}{}}{\Vector{a}}{\Vector{a}_{#1}}}\xspace}
\newcommand{\CONDNUM}{\ensuremath{\kappa}\xspace}

\newcommand{\CondNumk}[2]{\ensuremath{\CONDNUM(#1,#2)}\xspace}
\newcommand{\COVG}[3][]{\ensuremath{%
\ifthenelse{\equal{#1}{}}{G(#2,#3)}{G_{#1}(#2,#3)}}\xspace}
\newcommand{\COVGV}[2][]{\ensuremath{%
\ifthenelse{\equal{#1}{}}{\tilde{G}(#2)}{\tilde{G}_{#1}(#2)}}\xspace}

\newcommand{\remove}[1]{}

\newcommand{\vecvar}[1]{\ensuremath{\mathbf{#1}}\xspace}

\newcommand{\AGG}[1][]{\ensuremath{%
\ifthenelse{\equal{#1}{}}{g}{g_{#1}}}\xspace}

\newcommand{\SE}[1][]{\ensuremath{%
\ifthenelse{\equal{#1}{}}{\vecvar{x}}{\vecvar{x^{(#1)}}}}\xspace}

\newcommand{\Se}[2][]{\ensuremath{%
\ifthenelse{\equal{#1}{}}{X_{#2}}{X^{(#1)}_{#2}}}\xspace}

\newcommand{\ZERO}[1][]{\ensuremath{%
\ifthenelse{\equal{#1}{}}{\vecvar{0}}{\vecvar{0^{(#1)}}}}\xspace}

\newcommand{\ERRF}[1][]{\ensuremath{%
\ifthenelse{\equal{#1}{}}{g}{g_{#1}}}\xspace}
\newcommand{\ErrF}[2][]{\ensuremath{%
\ifthenelse{\equal{#1}{}}{g(#2)}{g_{#1}(#2)}}\xspace}

\begin{document}
\title{Submodular meets Spectral: Greedy Algorithms for 
Subset Selection, Sparse Approximation and Dictionary Selection}
\vspace{0.8cm}
\author{Abhimanyu Das\\
University of Southern California\\
{\tt abhimand@usc.edu}
\and
David Kempe\thanks{Supported in part by NSF CAREER award 0545855, and
  NSF grant DDDAS-TMRP 0540420}\\
University of Southern California\\
{\tt \dkempeemail}
}
\date{}
\maketitle

\begin{abstract}
We study the problem of selecting a subset of $k$ random
variables from a large set, in order to obtain the best
linear prediction of another variable of interest.
This problem can be viewed in the context of both feature
selection and sparse approximation.
We analyze the performance of widely used greedy
heuristics, using insights from the maximization of submodular
functions and spectral analysis.
We introduce the \todef{submodularity ratio} as a key quantity to
help understand why greedy algorithms 
perform well even when the variables are highly correlated. Using our
techniques, we obtain the strongest known approximation 
guarantees for this problem, both in terms of the submodularity
ratio and the smallest $k$-sparse eigenvalue of the covariance matrix.

We further demonstrate the wide applicability of our techniques by
analyzing greedy algorithms for the dictionary selection problem,
and significantly improve the previously known guarantees.
Our theoretical analysis is complemented by
experiments on real-world and synthetic data sets; 
the experiments show that the submodularity ratio is a stronger
predictor of the performance of greedy algorithms than other spectral
parameters.
\end{abstract}


\section{Introduction}

We analyze algorithms for the following important \todef{Subset
 Selection} problem: 
select a subset of $k$ variables from a given set of 
$n$ observation variables which, taken together, ``best''
predict another variable of interest.
This problem has many applications ranging from 
feature selection, sparse learning and dictionary selection 
in machine learning, to sparse approximation and 
compressed sensing in signal processing.
From a machine learning perspective, the variables could be features or 
observable attributes of a phenomenon, and we wish to 
predict the phenomenon using only a small subset from the
high-dimensional feature space.
In signal processing, the variables could correspond
to a collection of dictionary vectors, and the goal is to parsimoniously 
represent another (output) vector. 
For many practitioners, the prediction model of choice is 
linear regression, and
the goal is to obtain a linear model using a small subset of
variables, to minimize the mean square prediction error or,
equivalently, maximize the squared multiple correlation $R^2$ \cite{johnson}.

Thus, we formulate the Subset Selection problem for regression as follows:
Given the (normalized) covariances between 
$n$ variables $X_i$ (which can in principle be observed) and a 
variable $Z$ (which is to be predicted), select a subset of $k \ll n$
of the variables $X_i$ and a linear prediction function of $Z$ from
the selected $X_i$ that maximizes the $R^2$ fit.
(A formal definition is given in Section \ref{sec:preliminaries}.)
The covariances are usually obtained empirically from 
detailed past observations of the variable values.

The above formulation is known \cite{das:kempe} to be 
equivalent to 
the problem of \todef{sparse approximation} over dictionary vectors: 
the input consists of a dictionary of $n$ feature vectors 
$\Vector{x_i} \in \R^m$, along with a target vector 
$\Vector{z} \in \R^m$, and the goal is to select at most $k$ vectors
whose linear combination best approximates $\Vector{z}$. The
pairwise covariances of the previous formulation are then
exactly the inner products of the dictionary vectors.\footnote{
For this reason, the dimension $m$ of the feature vectors
only affects the problem indirectly, via the accuracy of the
estimated covariance matrix.}

Our problem formulation appears somewhat similar to the problem of sparse
recovery~\cite{zhangfb,zhangf,yu, candes}; however, note that in
sparse recovery, it is generally assumed that the 
prediction vector is truly (almost) $k$-sparse, and the aim is to
recover the exact coefficients of this truly sparse solution.
However, finding a sparse solution is a well-motivated problem
even if the true solution is not sparse.
Even then, running subset selection to find a sparse approximation to
the correct solution helps to reduce cost and model
complexity.

This problem is \NP-hard \cite{natarajan}, so 
no polynomial-time algorithms are known to solve it optimally for all inputs. 
Two approaches are frequently used for approximating such problems:
greedy algorithms \cite{miller,tropp3,muthu,zhangfb} and 
convex relaxation schemes \cite{tibshirani, candes, tropp4, donoho}. 
For our formulation, a disadvantage of convex relaxation techniques is
that they do not provide explicit control over the target sparsity
level $k$ of the solution; additional effort is needed to tune the 
regularization parameter. 

A simpler and more intuitive approach, widely used
in practice for subset selection problems (for example, it is
implemented in all commercial statistics packages), is to
use greedy algorithms, which iteratively add or remove 
variables based on simple measures of fit with $Z$. 
Two of the most well-known and widely used greedy algorithms
are the subject of our analysis:
Forward Regression \cite{miller} and Orthogonal Matching Pursuit 
\cite{tropp3}.
(These algorithms are defined formally in Section \ref{sec:preliminaries}).
   
So far, the theoretical bounds on such greedy 
algorithms have been unable to explain why they
perform well in practice for most subset selection problem instances.
Most previous results for greedy subset selection algorithms
\cite{muthu, tropp3, das:kempe} have been based on 
coherence of the input data, i.e., the maximum correlation $\mu$
between any pair of variables.
Small coherence is an extremely strong condition, and the
bounds 
usually break down when
the coherence is $\omega(1/k)$. 
On the other hand, most bounds for greedy and convex relaxation algorithms 
for sparse recovery are based on a weaker
sparse-eigenvalue or Restricted Isometry Property (RIP) condition
\cite{zhangf,zhangfb,lozano,zhou,candes}.
However, these results apply to a different 
objective: minimizing the difference between the actual and 
estimated coefficients of a sparse vector. Simply extending these 
results to the subset selection problem 
adds a 
dependence on the largest $k$-sparse eigenvalue and 
only leads to weak additive bounds. 
More importantly, all the above results rely on spectral
conditions that suffer from an inability to explain the
performance of the algorithms for near-singular matrices.

Eigenvalue-based bounds fail to explain an
observation of many experiments (including ours
in Section \ref{sec:experiments}): greedy algorithms often
perform very well, even for near-singular input matrices.
Our results begin to explain these observations by proving 
that the performance of many algorithms does not really depend on how
singular the covariance matrix is, but rather on how far the
$R^2$ measure deviates from submodularity on the given input.
We formalize this intuition by defining a measure of
``approximate submodularity'' which we term
\todef{submodularity ratio}.
We prove that whenever the submodularity ratio is bounded away from 0,
the $R^2$ objective is ``reasonably close'' to submodular, and
Forward Regression gives a constant-factor approximation.
This significantly generalizes a recent result of Das and Kempe
\cite{das:kempe}, who had identified a strong condition 
termed ``absence of conditional suppressors'' which ensures that 
the $R^2$ objective is actually submodular.

An analysis based on the submodularity ratio does relate with
traditional spectral bounds, in that the ratio is always lower-bounded
by the smallest $k$-sparse eigenvalue of \COVM (though it can be
significantly larger when the predictor variable is not badly
aligned with the eigenspace of small eigenvalues).
In particular, we also obtain multiplicative approximation guarantees for
both Forward Regression and Orthogonal Matching Pursuit, whenever the smallest
$k$-sparse eigenvalue of \COVM is bounded away from $0$, significantly 
strengthening past known bounds on their performance.

An added benefit of our framework is that we obtain much tighter
theoretical performance guarantees for greedy algorithms for
dictionary selection \cite{krause}. 
In the \todef{dictionary selection problem}, we are given $s$ target
vectors, and a candidate set $V$ of feature vectors. The goal is
to select a set $D \subset V$ of at most $d$ feature vectors,
which will serve as a \todef{dictionary} in the following sense.
For each of several target vectors, the best $k < d$ vectors from $D$
will be selected and used to achieve a good $R^2$ fit; the goal is
to maximize the average $R^2$ fit for all of these vectors.
(A formal definition is given in Section \ref{sec:preliminaries}.)
This problem of finding a dictionary of basis functions 
for sparse representation of signals has several applications in machine
learning and signal processing. Krause and Cevher \cite{krause} showed
that greedy algorithms for dictionary selection can perform well in many
instances, and proved additive approximation bounds for two specific
algorithms, \SDSMA and \SDSOMP (defined in Section \ref{sec:dictionaryselection}).
Our approximate submodularity framework allows us to obtain much
stronger multiplicative guarantees without much extra effort.

Our theoretical analysis is complemented by experiments
comparing the performance of the greedy algorithms and a 
baseline convex-relaxation algorithm for subset selection on two
real-world data sets and a synthetic data set.
More importantly, we evaluate the submodularity ratio of these
data sets and compare it with other spectral parameters: while 
the input covariance matrices are close to singular,
the submodularity ratio actually turns out to be significantly larger. 
Thus, our theoretical results can begin to explain why, in
many instances, greedy algorithms perform well in spite of the 
fact that the data may have high correlations.

Our main contributions can be summarized as follows:
\begin{enumerate}
\item We introduce the notion of the submodularity ratio as 
a much more accurate predictor of the performance of greedy algorithms
than previously used parameters.
\item We obtain the strongest known theoretical performance guarantees
for greedy algorithms for subset selection. 
In particular, we show (in Section \ref{sec:subsetselection}) that the
Forward Regression and OMP algorithms are within a 
$1-e^{-\Superm}$ factor and $1 - e^{-(\Superm \cdot \Eig_{\min})}$ factor
of the optimal solution, respectively (where the \Superm and $\Eig$
terms are appropriate submodularity and sparse-eigenvalue parameters).

\item We obtain the strongest known theoretical guarantees
for algorithms for dictionary selection, improving on the results of
\cite{krause}.
In particular, we show (in Section \ref{sec:dictionaryselection}) that
the \SDSMA algorithm is within a factor
$\frac{\Superm}{\Eig_{\max}}(1 -\frac{1}{e})$ of optimal.

\item
We evaluate our theoretical bounds for subset selection 
by running greedy and L1-relaxation algorithms on real-world and
synthetic data, and show how the various submodular and spectral
parameters correlate with the performance of the algorithms in
practice.

\end{enumerate}


\subsection{Related Work}

As mentioned previously, there has been a lot of recent interest in
greedy and convex relaxation techniques for the sparse recovery problems, both
in the noiseless and noisy setting. For L1 relaxation techniques, 
Tropp \cite{tropp4} showed conditions based on the coherence (i.e., the
maximum correlation between any pair of variables) of the
dictionary that guaranteed near-optimal recovery of a sparse signal. 
In \cite{candes,donoho}, it was shown that if the target signal
is truly sparse, and the dictionary obeys a restricted isometry property 
(RIP), then L1 relaxation can almost exactly recover the true sparse
signal.
Other results \cite{yu, zhou} also prove conditions
under which L1 relaxation can recover a sparse signal.
Though related, the above results are not directly applicable to our 
subset selection formulation, since the goal in sparse recovery is 
to recover the true coefficients of the sparse signal, as opposed to our
problem of minimizing the prediction error of an arbitrary signal 
subject to a specified sparsity level.

For greedy sparse recovery, Zhang \cite{zhangfb, zhangf} and 
Lozano et al.~\cite{lozano} provided conditions based on sparse eigenvalues
under which Forward Regression and Forward-Backward Regression 
can recover a sparse signal. As with the L1 results for 
sparse recovery,
the objective function analyzed in these papers is somewhat different
from that in our subset selection formulation; furthermore, these 
results are intended mainly for the case when the predictor variable 
is truly sparse. 
Simply extending these results to our problem formulation 
gives weaker, additive bounds and requires stronger conditions 
than our results.

The papers by Das and Kempe \cite{das:kempe}, Gilbert et
al.~\cite{muthu} and Tropp et al.~\cite{tropp2, tropp3} analyzed
greedy algorithms using the same subset selection formulation
presented in this work. In particular, they obtained 
a $1 + \Theta(\mu^2 k)$ multiplicative approximation guarantee for the mean square
error objective and a  $1 - \Theta(\mu k)$ guarantee for the $R^2$ objective,
whenever the coherence $\mu$ of the dictionary is $O(1/k)$. These results 
are thus weaker than those presented here, since they do not 
apply to instances with even moderate correlations of $\omega(1/k)$.

Other analysis of greedy methods includes the work of
Natarajan \cite{natarajan}, which proved  
a bicriteria approximation bound for minimizing the 
number of vectors needed to achieve a given prediction error. 

As mentioned earlier, the paper by Krause and Cevher \cite{krause}
analyzed greedy algorithms for the dictionary selection problem, which
generalizes subset selection to prediction of multiple variables. 
They too use a notion of approximate submodularity to provide 
additive approximation guarantees. Since their analysis is for a
more general problem than subset selection, applying their results 
directly to the subset selection problem predictably gives much weaker
bounds than those presented in this paper for subset selection. 
Furthermore, even for the general dictionary selection problem, 
our techniques can be used to 
significantly improve their  
analysis of greedy algorithms and obtain
tighter multiplicative approximation bounds (as shown in Section \ref{sec:dictionaryselection}).

In general, we note that the performance bounds for
greedy algorithms derived 
using the coherence parameter 
 are usually the weakest, followed by those using the Restricted
Isometry Property, then those using sparse eigenvalues, and finally
those using the submodularity ratio.
(We show an empirical comparison of these parameters in 
Section \ref{sec:experiments}.)


\section{Preliminaries} \label{sec:preliminaries}
The goal in subset selection is to estimate 
a \todef{predictor variable} $Z$ using linear
regression on a small subset from the set of \todef{observation variables} $V = \SET{X_1, \ldots, X_n}$. 
We use $\Var{X_i}$, $\Cov{X_i}{X_j}$ and $\Corr{X_i}{X_j}$ to denote
the variance, covariance and correlation of random variables, 
respectively.
By appropriate normalization, we can assume 
that all the random variables have expectation 0 and variance 1.
The matrix of covariances between the $X_i$ and $X_j$ is denoted by \COVM,
with entries  
$\CovM{i}{j} = \Cov{X_i}{X_j}$.
Similarly, we use \COVV to denote the covariances between $Z$ and the $X_i$,
with entries $\CovV{i} = \Cov{Z}{X_i}$.
Formally, the \todef{Subset Selection} problem can now be stated as
follows: 

\begin{definition}[Subset Selection] \label{def:subset-selection}
Given pairwise covariances among all variables, as well as a parameter $k$,
find a set $S \subset V$ of at most $k$ variables $X_i$
and a linear predictor $Z' = \sum_{i \in S} \Coeff{i} X_i$ of $Z$,
maximizing the \todef{squared multiple correlation} \cite{diekhoff,johnson}
\begin{align*}
\Rt{Z}{S} 
& = \frac{\Var{Z} - \Expect{(Z-Z')^2}}{\Var{Z}}.
\end{align*}
\end{definition}

$R^2$ is a widely used measure for the goodness of a
statistical fit; it captures the fraction of the variance of $Z$
explained by variables in $S$. 
Because we assumed $Z$ to be normalized to have variance 1, it
simplifies to $\Rt{Z}{S} = 1 - \Expect{(Z-Z')^2}$.

For a set $S$, we use \COVM[S] to denote the submatrix of \COVM
with row and column set $S$, and \COVV[S] to denote the vector with
only entries \CovV{i} for $i \in S$.
For notational convenience, we frequently do not distinguish between
the index set $S$ and the variables \Set{X_i}{i \in S}.
Given the subset $S$ of variables used for prediction,
the optimal regression coefficients \Coeff{i} are well known to be
$\COEFFVEC[S] = (\Coeff{i})_{i \in S} = \Inverse{\COVM[S]} \cdot \COVV[S]$
(see, e.g., \cite{johnson}), and hence 
$\Rt{Z}{S} = \Transpose{\COVV[S]} \Inverse{\COVM[S]} \COVV[S]$.
Thus, the subset selection problem can be phrased as follows:
Given \COVM, \COVV, and $k$, select a set $S$ of at most $k$ variables
to maximize
$\Rt{Z}{S} = \Transpose{\COVV[S]} (\Inverse{\COVM[S]})
\COVV[S]$.\footnote{We assume throughout that \COVM[S] is
  non-singular. For some of our results, an extension to singular
  matrices is possible using the Moore-Penrose generalized inverse.}

The dictionary selection problem generalizes the subset selection problem by 
considering $s$ predictor variables $Z_1, Z_2,\ldots, Z_s$. The goal is
to select a dictionary $D$ of $d$ observation variables, to optimize
the average $R^2$ fit for the $Z_i$ using at most $k$ vectors from $D$
for each. Formally, the Dictionary Selection problem is
defined as follows: 

\begin{definition}[Dictionary Selection] \label{def:dictionary-selection}
Given all pairwise covariances among the $Z_j$ and $X_i$ variables, 
as well as parameters $d$ and $k$,
find a set $D$ of at most $d$ variables from $\SET{X_1,\ldots, X_n}$ 
 maximizing
\begin{align*}
\FD{D} & = \sum_{j=1}^s \max_{S \subset D,\SetCard{S}=k}\Rt{Z_j}{S}.
\end{align*}
\end{definition}

Many of our results are phrased in terms of eigenvalues of the
covariance matrix \COVM and its submatrices. 
Since covariance matrices are positive semidefinite,
their eigenvalues are real and non-negative \cite{johnson}.
For any positive semidefinite $n \times n$ matrix $A$, we
denote its eigenvalues by
$\eigmin{A} = \lambda_1(A) \leq \lambda_2(A) \leq \ldots \leq 
\lambda_n(A) = \eigmax{A}$.
We use $\eigmink{\COVM}{k} = \min_{S: \SetCard{S}=k} \eigmin{\COVM[S]}$ to
refer to the smallest eigenvalue of any $k \times k$ submatrix of
$\COVM$ (i.e., the smallest $k$-sparse eigenvalue), and similarly 
$\eigmaxk{\COVM}{k} = \max_{S: \SetCard{S}=k} \eigmax{\COVM[S]}$.
\footnote{
Computing $\eigmink{\COVM}{k}$ is NP-hard. In Appendix  
\ref{app:eigptas} we describe how to efficiently approximate 
the values for some scenarios.
}
We also use \CondNumk{C}{k} to denote the largest condition number
(the ratio of the largest and smallest eigenvalue) 
of any $k \times k$ submatrix of $\COVM$. This quantity is strongly related
to the Restricted Isometry Property in \cite{candes}. We also use 
$\mu(\COVM)=\max_{i \neq j} \Abs{\CovM{i}{j}}$ to denote the 
\todef{coherence}, i.e., the maximum absolute pairwise 
correlation between the $X_i$ variables.
Recall the $L_2$ vector and matrix norms:
$\Norm[2]{\Vector{x}} = \sqrt{\sum_i \Abs{x_i}^2}$, and
$\Norm[2]{A} = \eigmax{A} = \max_{\Norm[2]{\Vector{x}} = 1} \Norm[2]{A\Vector{x}}$.
We also use 
$\Norm[0]{\Vector{x}} = \SetCard{\Set{i}{x_i \neq 0}}$ to denote the
sparsity of a vector \Vector{x}. 

\Omit{The latter (Rayleigh-Ritz) representation is also useful in bounding
$\eigmin{A}$, as for any vector $\Vector{x}$, we have
$\eigmin{A} \leq \frac{\Norm[2]{A\Vector{x}}}{\Norm[2]{\Vector{x}}}$.
}

The part of a variable $Z$ that is not correlated with the $X_i$ for
all $i \in S$, i.e., the part that cannot be explained by the $X_i$,
is called the \todef{residual} (see \cite{diekhoff}), and defined as 
$\Residual{Z}{S} = Z-\sum_{i \in S} \Coeff{i} X_i$. 

\subsection{Submodularity Ratio}
\remove{Our main results in this paper are multiplicative
  approximation guarantees for greedy algorithms in terms of a
  quantity that we define as the submodularity ratio.}

We introduce the notion of submodularity ratio for a general set
function, which captures ``how close'' to submodular the function is.
We first define it for arbitrary set functions, and then show the
specialization for the $R^2$ objective.

\begin{definition}[Submodularity Ratio]
\label{def:supermod}
Let $f$ be a non-negative set function.
The \todef{submodularity ratio} of $f$ with respect to a set
$U$ and a parameter $k \geq 1$ is
\begin{align*}
\superm[f]{U}{k} 
& = \min_{L \subseteq U, S : \SetCard{S} \leq k, S \cap L = \emptyset}
\frac{\sum_{x \in S} f(L \cup \SET{x}) - f(L)}{f(L \cup S) - f(L)}.
\end{align*}
Thus, it captures how much more $f$ can increase by adding any
subset $S$ of size $k$ to $L$, compared to the combined benefits of 
adding its individual elements to $L$.
\item If $f$ is specifically the $R^2$ objective defined on the
variables $X_i$, then we omit $f$ and simply define
\[
\superm{U}{k} 
\; = \; \min_{L \subseteq U, S: \SetCard{S} \leq k, S \cap L = \emptyset}
\frac{\sum_{i \in S} (\Rt{Z}{L \cup \SET{X_i}} - \Rt{Z}{L})}{%
\Rt{Z}{S \cup L} - \Rt{Z}{L}} 
\; = \; \min_{L \subseteq U, S: \SetCard{S} \leq k, S \cap L = \emptyset}
\frac{\Transpose{(\COVVS{L}_S)} \COVVS{L}_S}{{\Transpose{(\COVVS{L}_S)}} 
\Inverse{(\COVMS{L}_S)} \COVVS{L}_S},
\]
where \COVMS{L} and \COVVS{L} 
are the normalized covariance matrix and normalized covariance vector 
corresponding to the set
\SET{\Residual{X_1}{L}, \Residual{X_2}{L},\ldots, \Residual{X_n}{L}}.
\end{definition}

It can be easily shown that $f$ is submodular if and only if
$\superm{U}{k} \geq 1$, for all $U$ and $k$. 
For the purpose of subset selection, it is significant
that the submodularity ratio can be bounded in terms of the smallest 
sparse eigenvalue, as shown in the following lemma:

\begin{lemma} \label{lem:supermod}
$\superm{U}{k} \geq \eigmink{\COVM}{k + \SetCard{U}} \geq \eigmin{\COVM}$.
\end{lemma}

For all our analysis in this paper, 
we will use $\SetCard{U}=k$, and hence $\superm{U}{k} \geq \eigmink{\COVM}{2k}$.
Thus, the smallest $2k$-sparse eigenvalue is a lower bound on this submodularity
ratio; as we show later, it is often a weak lower bound.

Before proving Lemma \ref{lem:supermod},  we first introduce two lemmas 
that relate the eigenvalues of normalized covariance matrices with those of its submatrices.

\begin{lemma} \label{lem:correig}
Let \COVM be the covariance matrix of $n$ zero-mean random variables
$X_1,X_2,\ldots, X_n$, each of which has variance at most $1$.
Let \COVMC be the corresponding correlation matrix of the $n$ random
variables, that is, $\COVMC$ is the covariance matrix of the
variables after they are normalized to have unit variance.
Then $\eigmin{\COVM} \leq \eigmin{\COVMC}$.
\end{lemma}

\begin{proof}
Since \COVMC is obtained by normalizing the variables such that they have
unit variance, we get
$\COVMC = D^T \COVM D$, where $D$ is a diagonal matrix with
diagonal entries $d_i = \frac{1}{\sqrt{\Var{X_i}}}$.

Since both \COVMC and \COVM are positive semidefinite, we can perform
 Cholesky factorization to get lower-triangular matrices \VMC and \VM
such that $\COVM = \VM \VM^T$ and $\COVMC = \VMC \VMC^T$. Hence
$\VMC = D^T \VM$.

Let \singmin{\VM} and \singmin{\VMC}denote the smallest singular values
of \VM and \VMC, respectively.
Also, let \Vector{v} be the singular vector corresponding to
\singmin{\VMC}. Then,
\[
\Norm[2]{\VM \Vector{v}}
\; = \; \Norm[2]{\Inverse{D} \VMC \Vector{v}}
\; \leq \; \Norm[2]{\Inverse{D}}\Norm[2]{\VMC \Vector{v}}
\; = \; \singmin{\VMC} \Norm[2]{\Inverse{D}}
\; \leq \; \singmin{\VMC},
\]
where the last inequality follows since 
\[
\Norm[2]{\Inverse{D}} 
\; = \; \max_i \frac{1}{d_i} 
\; = \; \max_i \sqrt{\Var{X_i}}
\; \leq \; 1.
\]

Hence, by the Courant-Fischer
theorem, $\singmin{\VM} \leq \singmin{\VMC}$, and consequently,
$\eigmin{\COVM} \leq \eigmin{\COVMC}$.
\end{proof}

\begin{lemma} \label{lem:condeig}
Let \eigmin{\COVM} be the smallest eigenvalue of the
covariance matrix $\COVM$ of $n$ random variables
$X_1,X_2,\ldots, X_n$, and \eigmin{\COVMP} be the smallest
eigenvalue of the $(n-1) \times (n-1)$ covariance matrix \COVMP
corresponding to the $n-1$ random variables
$\Residual{X_1}{X_n}, \ldots, \Residual{X_{n-1}}{X_n}$.
Then $\eigmin{\COVM} \leq \eigmin{\COVMP}$.
\end{lemma}

\begin{proof}
Let $\lambda_i$  and $\lambda'_i$ denote the eigenvalues of \COVM
and \COVMP respectively. Also, let \CovMP{i}{j} denote the entries of \COVMP.
Using the definition of the residual, we get that
\[ \begin{array}{lclcl}
\CovMP{i}{j} 
& = & \Cov{\Residual{X_{i}}{X_n}}{\Residual{X_{j}}{X_n}} 
& = & \CovM{i}{j} - \frac{\CovM{i}{n}\CovM{j}{n}}{\CovM{n}{n}},\\
\CovMP{i}{i} 
& = & \Var{\Residual{X_{i}}{X_n}} 
& = & \CovM{i}{i} - \frac{\CovM{i}{n}^2}{\CovM{n}{n}}.
\end{array} \]
Defining $D = \frac{1}{\CovM{n}{n}}
\cdot \Transpose{[\CovM{1}{n}, \CovM{2}{n}, \ldots,\CovM{n-1}{n}]}
\cdot [\CovM{1}{n}, \CovM{2}{n}, \ldots, \CovM{n-1}{n}]$,
we can write $\COVM[\SET{1, \ldots, n-1}] = \COVMP + D$.
To prove $\lambda_1 \leq \lambda'_1$,
let $\Vector{e'} = \Transpose{[e'_1,\ldots,e'_{n-1}]}$ be
the eigenvector of \COVMP corresponding to the eigenvalue $\lambda'_1$,
and consider the vector
$\mathbf{e}=\Transpose{[e'_1, e'_2,\ldots,e'_{n-1}, -\frac{1}{\CovM{n}{n}} \sum_{i=1}^{n-1} e'_{i}\CovM{i}{n}]}$.
Then, $\COVM \cdot \Vector{e} =
\left[\begin{smallmatrix}\Vector{y} \\ 0 \end{smallmatrix} \right]$, where
\begin{align*}
\Vector{y} & = 
-\frac{1}{\CovM{n}{n}} \sum_{i=1}^{n-1} e'_{i}\CovM{i}{n}
\Transpose{[\CovM{1}{n},\CovM{2}{n}, \ldots, \CovM{n-1}{n}]} 
+ \COVM[\SET{1, \ldots, n-1}] \cdot \Vector{e'} \\
& = -\frac{1}{\CovM{n}{n}} \sum_{i=1}^{n-1} e'_{i}\CovM{i}{n}
\Transpose{[\CovM{1}{n},\CovM{2}{n}, \ldots, \CovM{n-1}{n}]} 
+ D \cdot \Vector{e'} + \COVMP \cdot \Vector{e'} \\
& = \COVMP \cdot \Vector{e'}.
\end{align*}

Thus,
$\COVM \cdot \Vector{e} =
\Transpose{[\lambda'_1 e'_1,  \lambda'_1 e'_2, \ldots, \lambda'_1 e'_{n-1}, 0]}
= \lambda'_1 \Transpose{[e'_1, e'_2,\ldots,e'_{n-1}, 0]}
\leq \lambda'_1 \Norm[2]{\Vector{e}}$,
which by Rayleigh-Ritz bounds implies that $\lambda_1 \leq \lambda'_1$.
\end{proof}

Using the above two lemmas, we now prove Lemma \ref{lem:supermod}.
\begin{extraproof}{Lemma \ref{lem:supermod}}
Since
\[ 
\frac{{\Transpose{(\COVVS{L}_S)}}
\Inverse{(\COVMS{L}_S)} \COVVS{L}_S}{\Transpose{(\COVVS{L}_S)} \COVVS{L}_S}
\; \leq \; \max_{\Vector{x}}
\frac{\Transpose{\Vector{x}}
\Inverse{(\COVMS{L}_S)} \Vector{x}}{{\Transpose{\Vector{x}} \Vector{x}}} 
\; = \; \eigmax{\Inverse{(\COVMS{L}_S)}} 
\; = \; \frac{1}{\eigmin{\COVMS{L}_S}},
\]
we can use Definition \ref{def:supermod} to obtain that
\begin{align*}
\superm{U}{k} 
& \geq \min_{(L \subseteq U, S: \SetCard{S} \leq k, S \cap L = \emptyset)} \eigmin{\COVMS{L}_S}.
\end{align*}

Next, we relate $\eigmin{\COVMS{L}_S}$ with $\eigmin{\COVM_{L \cup S}}$,
using repeated applications of Lemmas \ref{lem:correig} and \ref{lem:condeig}.
Let $L = \SET{X_1, \ldots, X_\ell}$; for each $i$, define
$L_i = \SET{X_1, \ldots, X_i}$, and let
$\COVM^{(i)}$ be the covariance matrix of the random variables
$\Set{\Residual{X}{L \setminus L_i}}{X \in S \cup L_i}$, and
$\COVM^{(i)}_\rho$ the covariance matrix after normalizing all its variables
to unit variance.
Then, Lemma \ref{lem:correig} implies that for each $i$,
$\eigmin{\COVM^{(i)}} \leq \eigmin{\COVM^{(i)}_\rho}$, and
Lemma \ref{lem:condeig} shows that
$\eigmin{\COVM^{(i)}_\rho} \leq \eigmin{\COVM^{(i-1)}}$
for each $i > 0$. Combining these inequalities inductively for all
$i$, we obtain that
\[
\eigmin{\COVMS{L}_S} 
\; = \; \eigmin{\COVM^{(0)}_\rho}
\; \geq \; \eigmin{\COVM^{(\ell)}}
\; = \; \eigmin{\COVM_{L \cup S}}
\; \geq \; \eigmink{\COVM}{\SetCard{L \cup S}}.
\]
Finally, since $\SetCard{S} \leq k$ and $L \subseteq U$,
we obtain $\superm{U}{k} \geq \eigmink{\COVM}{k + \SetCard{U}}$.
\end{extraproof}

\section{Algorithms Analysis}\label{sec:subsetselection}

We now present theoretical performance bounds for Forward
Regression and Orthogonal Matching Pursuit, which are widely
used in practice. We also analyze the Oblivious
algorithm: one of the simplest greedy algorithms for subset 
selection. 
Throughout this section, we use
$\OPT = \max_{S: \SetCard{S} = k} \Rt{Z}{S}$ to denote the optimum
$R^2$ value achievable by any set of size $k$. 

\subsection{Forward Regression} \label{sec:forward}

We first provide approximation bounds for Forward Regression,
which is the standard algorithm used by many researchers in
medical, social, and economic domains.\footnote{There is some
inconsistency in the literature about naming of greedy
algorithms. Forward Regression is sometimes also referred to as
Orthogonal Matching Pursuit (OMP). We choose the 
nomenclature consistent with \cite{miller} and \cite{tropp3}.}

\begin{definition}[Forward Regression] \label{def:forward-regression}
The \todef{Forward Regression} (also called \todef{Forward Selection})
algorithm for subset selection selects a set $S$ of size $k$
iteratively as follows: 
\begin{algorithmic}[1]
\STATE Initialize $S_0 = \emptyset$. 
\FOR{each iteration $i+1$}
\STATE Let $X_m$ be a variable maximizing $\Rt{Z}{S_i \cup \SET{X_m}}$,
and set $S_{i+1} = S_i \cup \SET{X_m}$.
\ENDFOR
\STATE Output $S_k$.
\end{algorithmic}
\end{definition}

Our main result is the following theorem.

\begin{theorem} \label{thm:frapprox}
The set $\FrSet$ selected by forward regression has the following
approximation guarantees:
\begin{align*}
\Rt{Z}{\FrSet} & \geq (1 - e^{-\superm{\FrSet}{k}}) \cdot \OPT\\
& \geq (1 - e^{-\eigmink{\COVM}{2k}}) \cdot \OPT\\
& \geq (1 - e^{-\eigmink{\COVM}{k}}) \cdot
\Theta((\half)^{1/\eigmink{\COVM}{k}}) \cdot \OPT.
\end{align*}
\end{theorem}

Before proving the theorem, we first begin with a general lemma 
that bounds the amount by which the $R^2$ value of a set 
and the sum of $R^2$ values of its elements can differ.

\begin{lemma} \label{lem:r2bounds}
\Omit{
Let \COVM and \COVV be the covariance matrix and covariance vector 
corresponding to a predictor variable $Z$ and a set $S$ of   
random variables $X_1,X_2,\ldots, X_n$ that are normalized to have zero mean
and unit variance.}
$\frac{1}{\eigmax{\COVM}} \sum_{i=1}^n \Rt{Z}{X_i} \leq
\Rt{Z}{\SET{X_1, \ldots, X_n}} 
\leq \frac{1}{\superm{\emptyset}{n}} \sum_{i=1}^n \Rt{Z}{X_i} \leq
\frac{1}{\eigmin{\COVM}} \sum_{i=1}^n \Rt{Z}{X_i}$.
\end{lemma}

\begin{proof}
Let the eigenvalues of $\COVM^{-1}$ be 
$\lambda'_1 \leq \lambda'_2 \leq \ldots \leq \lambda'_n$,
with corresponding orthonormal eigenvectors 
$\Vector{e_1}, \Vector{e_2}, \ldots, \Vector{e_n}$. 
We write \COVV in the basis 
$\SET{\Vector{e_1}, \Vector{e_2}, \ldots, \Vector{e_n}}$
as $\COVV = \sum_i \beta_i \Vector{e_i}$. Then, 
\[ 
\Rt{Z}{\SET{X_1, \ldots, X_n}} 
\; = \; \Transpose{\COVV} \COVM^{-1} \COVV 
\; = \; \sum_i \beta_i^2 \lambda'_i.
\]
Because $\lambda'_1 \leq \lambda'_i$ for all $i$, we get
$\lambda'_1 \sum_i \beta_i^2 \leq \Rt{Z}{\SET{X_1, \ldots, X_n}}$,
and $\sum_i \beta_i^2 = \Transpose{\COVV} \COVV = \sum_i \Rt{Z}{X_i}$,
because the length of the vector \COVV is independent of the basis it
is written in. Also, by definition of the submodularity ratio,
$\Rt{Z}{\SET{X_1, \ldots, X_n}} \leq \frac{\sum_i \beta_i^2}{\superm{\emptyset}{n}}$.
Finally, because $\lambda'_1 = \frac{1}{\eigmax{\COVM}}$, and using 
Lemma \ref{lem:supermod}, we obtain the result.
\end{proof}

The next lemma relates the optimal $R^2$ value using $k$ elements 
to the optimal $R^2$ using $k' < k$ elements.

\begin{lemma} \label{lem:optk}
For each $k$, let $\OPTSET{k} \in \argmax_{\SetCard{S} \leq k} \Rt{Z}{S}$
be an optimal subset of at most $k$ variables.
Then, for any $k' = \Theta(k)$ such that 
$\frac{1}{\eigmink{\COVM}{k}} < k' < k$, 
we have that
$\Rt{Z}{\OPTSET{k'}} \geq 
\Rt{Z}{\OPTSET{k}} \cdot \Theta((\frac{k'}{k})^{1/\eigmink{\COVM}{k}})$,
 for large enough $k$.
In particular,
$\Rt{Z}{\OPTSET{k/2}} \geq 
\Rt{Z}{\OPTSET{k}} \cdot \Theta((\half)^{1/\eigmink{\COVM}{k}})$, for
large enough $k$.
\end{lemma}

\begin{proof}
We first prove that
$\Rt{Z}{\OPTSET{k-1}} \geq (1 - \frac{1}{k \eigmink{\COVM}{k}}) \Rt{Z}{\OPTSET{k}}$.
Let $T = \Residual{Z}{\OPTSET{k}}$; then, $\Cov{X_i}{T} = 0$ for all
$X_i \in \OPTSET{k}$, and $Z = T + \sum_{X_i \in \OPTSET{k}} \alpha_i X_i$,
where $\Vector{\alpha} = (\alpha_i) = \Inverse{\COVM[\OPTSET{k}]} \cdot \COVV[\OPTSET{k}]$
are the optimal regression coefficients.
We write $Z' = Z - T$.
For any $X_j \in \OPTSET{k}$, by definition of $R^2$, we have that
\[ 
\Rt{Z'}{\OPTSET{k} \setminus \SET{X_j}} 
\; = \; 1 - \frac{\alpha_j^2 \Var{X_j}}{\Var{Z'}}
\; = \; 1 - \frac{\alpha_j^2}{\Var{Z'}};
\]
in particular, this implies that
$\Rt{Z'}{\OPTSET{k-1}} \geq 1 - \frac{\alpha_j^2}{\Var{Z'}}$ for all
$X_j \in \OPTSET{k}$.

Focus now on $j$ minimizing $\alpha_j^2$, so that
$\alpha_j^2 \leq  \frac{\Norm[2]{\Vector{\alpha}}^2}{k}$.
As in the proof of Lemma \ref{lem:r2bounds}, by
writing $\Vector{\alpha}$ in terms of an
orthonormal eigenbasis of \COVM[\OPTSET{k}], one can show that
$\Abs{\Transpose{\Vector{\alpha}} \COVM[\OPTSET{k}] \Vector{\alpha}}
\geq \Norm[2]{\Vector{\alpha}}^2 \eigmin{\COVM[\OPTSET{k}]}$, or
$\Norm[2]{\Vector{\alpha}}^2 \leq
\frac{\Abs{\Transpose{\Vector{\alpha}} \COVM[\OPTSET{k}] \Vector{\alpha}}}{\eigmin{\COVM[\OPTSET{k}]}}$.
Furthermore, $\Transpose{\Vector{\alpha}} \COVM[\OPTSET{k}] \Vector{\alpha}
= \Var{\sum_{X_i \in \OPTSET{k}} \alpha_i X_i} = \Var{Z'}$, so
$\Rt{Z'}{\OPTSET{k-1}} \geq 1 - \frac{1}{k \eigmin{\COVM[\OPTSET{k}]}}$.
Finally, by definition, $\Rt{Z'}{\OPTSET{k}} = 1$, so
\[ 
\frac{\Rt{Z}{\OPTSET{k-1}}}{\Rt{Z}{\OPTSET{k}}}
\; \geq \; \frac{\Rt{Z'}{\OPTSET{k-1}}}{\Rt{Z'}{\OPTSET{k}}}
\; \geq \; 1 - \frac{1}{k \eigmin{\COVM[\OPTSET{k}]}}
\; \geq \; 1 - \frac{1}{k \eigmink{\COVM}{k}}.
\]

Now, applying this inequality repeatedly, we get
\begin{align*}
\Rt{Z}{\OPTSET{k'}} 
& \geq \Rt{Z}{\OPTSET{k}} \cdot
\prod_{i=k'+1}^k (1 - \frac{1}{i \eigmink{\COVM}{i}}).
\end{align*}


Let $t=\Ceiling{1/\eigmink{\COVM}{k}}$, so that the previous bound implies
$\Rt{Z}{\OPTSET{k'}} \geq \Rt{Z}{\OPTSET{k}} \cdot  \prod_{i=k'+1}^k \frac{i - t}{i}$.
Most of the terms in the product telescope, giving us a bound of
$\Rt{Z}{\OPTSET{k}} \cdot \prod_{i=1}^{t} \frac{k' - t + i}{k - t + i}$.
Since $\prod_{i=1}^t \frac{k' - t + i}{k - t + i}$ converges to
$(\frac{k'}{k})^t$ with increasing $k$ (keeping $t$ constant),
we get that for large $k$,
\[
\Rt{Z}{\OPTSET{k'}}
\; \geq \; \Rt{Z}{\OPTSET{k}} \cdot \Theta((\frac{k'}{k})^t)
\; \geq \; \Rt{Z}{\OPTSET{k}} \cdot \Theta((\frac{k'}{k})^{1/\eigmink{\COVM}{k}}).
\]
\end{proof}

Using the above lemmas, we now prove the main theorem.

\begin{extraproof}{Theorem \ref{thm:frapprox}}
We begin by proving the first inequality.
Let $\OPTSET{k}$ be the optimum set of variables.
Let $\GrSet{i}$ be the set of variables chosen by Forward Regression
in the first $i$ iterations, and $S_i = \OPTSET{k} \setminus \GrSet{i}$.
By monotonicity of $R^2$ and the fact that
$S_i \cup \GrSet{i} \supseteq \OPTSET{k}$, we have that
$\Rt{Z}{S_i \cup \GrSet{i}} \geq \OPT$.

For each $X_j \in S_i$, let $X'_j = \Residual{X_j}{\GrSet{i}}$ be the residual of $X_j$
conditioned on $\GrSet{i}$, and write $S'_i = \Set{X'_j}{X_j \in S}$.

We will show that at least one of the $X'_i$ is a good candidate in
iteration $i+1$ of Forward Regression. First, the joint contribution
of $S'_i$ must be fairly large:
$\Rt{Z}{\Residual{S'_i}{\GrSet{i}}} = \Rt{Z}{S'_i} \geq \OPT - \Rt{Z}{\GrSet{i}}$.
Using Definition \ref{def:supermod}, 
as well as $\GrSet{i} \subseteq \FrSet$ and $\SetCard{S_i} \leq k$,
\[
\sum_{X'_j \in S'_i} \Rt{Z}{X'_j} 
\; \geq \; \superm{\GrSet{i}}{\SetCard{S_i}} \cdot \Rt{Z}{S'_i} 
\; \geq \; \superm{\FrSet}{k} \cdot \Rt{Z}{S'_i}.
\]
Let $\ell$ maximize $\Rt{Z}{X'_\ell}$, i.e., 
$\ell \in \argmax_{(j:X'_j \in S'_i)}\Rt{Z}{X'_j}$.
Then we get that
\[
\Rt{Z}{X'_\ell} 
\; \geq \; \frac{\superm{\FrSet}{k}}{\SetCard{S'_i}} \cdot \Rt{Z}{S'_i} 
\; \geq \; \frac{\superm{\FrSet}{k}}{k} \cdot \Rt{Z}{S'_i}.
\]

Define $A(i) = \Rt{Z}{\GrSet{i}} - \Rt{Z}{S_G^{i-1}}$ to be the gain obtained
from the variable chosen by Forward Regression in iteration $i$.
Then $\Rt{Z}{\FrSet} = \sum_{i=1}^k A(i)$.
Since the $X'_\ell$ above was a candidate to be chosen in iteration $i+1$, 
and Forward Regression chose a variable $X_{m}$ such that
$\Rt{Z}{\Residual{X_{m}}{\GrSet{i}}} \geq \Rt{Z}{\Residual{X}{\GrSet{i}}}$ 
for all $X \notin \GrSet{i}$, we obtain that 
\[
A(i+1) \; \geq \; \frac{\superm{\FrSet}{k}}{k} \cdot \Rt{Z}{S'_i} 
\; \geq \; \frac{\superm{\FrSet}{k}}{k} (\OPT - \Rt{Z}{\GrSet{i}})
\; \geq \; \frac{\superm{\FrSet}{k}}{k} (\OPT - \sum_{j=1}^{i} A(j)).
\]

Since the above inequality holds for each iteration $i=1,2,\ldots, k$, a
simple inductive proof establishes the bound
$\OPT - \sum_{i=1}^{k} A(i) \leq \OPT \cdot (1 - \frac{\superm{\FrSet}{k}}{k})^k$.
\Omit{
Since the above inequality is valid for each iteration i=1,2,\ldots k, we have
\begin{eqnarray*}
\OPT - \sum_{j=1}^{k}A(j) & = & \OPT - \sum_{j=1}^{k-1}A(j) - A(k)\\
 & \leq & \OPT - \sum_{j=1}^{k-1}A(j) -  \frac{\superm{\FrSet}{k}}{k} 
(\OPT - \sum_{j=1}^{i}A(j))\\
 & \leq & (\OPT - \sum_{j=1}^{k-1}A(j))(1 - \frac{\superm{\FrSet}{k}}{k})\\
 & \leq & (\OPT - \sum_{j=1}^{k-2}A(j) - A(k-1))(1 - \frac{\superm{\FrSet}{k}}{k})\\
 & \leq & (\OPT - \sum_{j=1}^{k-2}A(j))(1 - \frac{\superm{\FrSet}{k}}{k})^2\\
 &\ldots&\\
 & \leq & \OPT(1 - \frac{\superm{\FrSet}{k}}{k})^k \leq 
\OPT(\frac{1}{e})^{\superm{\FrSet}{k}}
\end{eqnarray*}
}
Hence,
\[
\Rt{Z}{\FrSet} 
\; = \; \sum_{i=1}^k A(i) 
\; \geq \; \OPT - \OPT(1 - \frac{\superm{\FrSet}{k}}{k})^k
\; \geq \; \OPT \cdot (1- e^{-\superm{\FrSet}{k}}).
\]

The second inequality follows directly from Lemma \ref{lem:supermod},
and the fact that $\SetCard{\FrSet}=k$.
By applying the above result after $k/2$ iterations, we obtain 
$\Rt{Z}{\GrSet{k/2}} \geq (1 - e^{-\eigmink{\COVM}{k}}) \cdot \Rt{Z}{\OPTSET{k/2}}$.
Now, using Lemma \ref{lem:optk} and monotonicity of $R^2$, we get 
\[
\Rt{Z}{\GrSet{k}} 
\; \geq \; \Rt{Z}{\GrSet{k/2}}
\; \geq \; (1 - e^{-\eigmink{\COVM}{k}})\cdot \Theta((\half)^{1/\eigmink{\COVM}{k}}) \cdot \Rt{Z}{\OPTSET{k}},
\]
proving the third inequality.
\end{extraproof} 

\subsection{Orthogonal Matching Pursuit}

The second greedy algorithm we analyze is Orthogonal Matching Pursuit 
(OMP), frequently used in signal processing domains.
\begin{definition}[Orthogonal Matching Pursuit (OMP)] \label{def:omp}
The \todef{Orthogonal Matching Pursuit} algorithm for subset selection
selects a set $S$ of size $k$ iteratively as follows: 
\begin{algorithmic}[1]
\STATE Initialize $S_0 = \emptyset$. 
\FOR{each iteration $i+1$}
\STATE Let $X_m$ be a variable maximizing $\Abs{\Cov{\Residual{Z}{S_i}}{X_m}}$, 
and set $S_{i+1} = S_i \cup \SET{X_m}$.
\ENDFOR
\STATE Output $S_k$.
\end{algorithmic}
\end{definition}

By applying similar techniques as in the previous section, 
we can also obtain approximation bounds for OMP.
We start by proving the following lemma that lower-bounds the 
variance of the residual of a variable.

\begin{lemma}
\label{lem:condvar}
Let $A$ be the $(n+1) \times (n+1)$ covariance matrix of the
normalized variables $Z, X_1, X_2,\ldots, X_n$. Then 
$\Var{\Residual{Z}{\SET{X_1,\ldots,X_n}}} \geq \eigmin{A}$.
\end{lemma}

\begin{proof}
The matrix $A$ is of the form
$A = \left( \begin{array}{cc} 1 & \Transpose{\COVV}\\ \COVV & \COVM \end{array} \right)$.
We use $A[i,j]$ to denote the matrix obtained by removing
the \Kth{i} row and \Kth{j} column of $A$, and similarly for \COVM.
Recalling that the $(i,j)$ entry of $\Inverse{\COVM}$ is
$\frac{(-1)^{i+j} \det(\COVM{[i,j]})}{\det(\COVM)}$,
and developing the determinant of $A$ by the first row and column,
we can write
\begin{align*}
\det(A)
&  =  \sum_{j=1}^{n+1} (-1)^{1+j} a_{1,j} \det(A[1,j])\\
&  =  \det(C) + \sum_{j=1}^n (-1)^j \CovV{j} \det(A[1,j+1])\\
&  =  det(C) +
   \sum_{j=1}^n (-1)^j \CovV{j} \sum_{i=1}^n
(-1)^{i+1} \CovV{i} \det(\COVM{[i,j]})\\
& = \det(C) - \sum_{j=1}^n \sum_{i=1}^n (-1)^{i+j} \CovV{i} \CovV{j} \det(\COVM{[i,j]})\\
& = \det(C)(1 - \Transpose{\COVV} \Inverse{\COVM} \COVV).
\end{align*}

Therefore, using that $\Var{Z} = 1$,
\[
\Var{\Residual{Z}{\SET{X_1,\ldots,X_n}}}
\; = \; \Var{Z} - \Transpose{\COVV} \Inverse{\COVM} \COVV
\; = \; \frac{\det(A)}{\det(\COVM)}.
\]
Because $\det(A) = \prod_{i=1}^{n+1} \lambda_i^A$ and
$\det(\COVM) = \prod_{i=1}^{n} \lambda_i^{\COVM}$, and
$\lambda_1^A \leq \lambda_1^{\COVM} \leq \lambda_2^A \leq
\lambda_2^{\COVM} \leq \ldots \leq \lambda_{n+1}^A$ by the eigenvalue
interlacing theorem, we get that $\frac{\det(A)}{\det(\COVM)} \geq \lambda_1^A$,
proving the lemma.
\end{proof}

The above lemma, along with an analysis similar to the proof of
Theorem \ref{thm:frapprox}, can be used to prove the following approximation
bounds for OMP: 

\begin{theorem} \label{thm:ompapprox}
The set $\OmpSet$ selected by orthogonal matching pursuit has the following
approximation guarantees:
\begin{align*}
\Rt{Z}{\OmpSet} 
& \geq (1 - e^{-(\superm{\OmpSet}{k} \cdot \eigmink{\COVM}{2k})}) \cdot \OPT\\
& \geq (1 - e^{-\eigmink{\COVM}{2k}^2}) \cdot \OPT\\
& \geq (1 - e^{-\eigmink{\COVM}{k}^2}) \cdot \Theta((\frac{1}{2})^{1/\eigmink{\COVM}{k}}) \cdot \OPT.
\end{align*}
\end{theorem}

\begin{proof}
We begin by proving the first inequality. Using notation similar
to that in the proof of Theorem \ref{thm:frapprox}, we
let $\OPTSET{k}$ be the optimum set of $k$ variables,
$\GrSet{i}$ the set of variables chosen by OMP
in the first $i$ iterations, and $S_i = \OPTSET{k} \setminus \GrSet{i}$.
For each $X_j \in S_i$, let $X'_j = \Residual{X_j}{\GrSet{i}}$ be the residual of $X_j$
conditioned on $\GrSet{i}$, and write $S'_i = \Set{X'_j}{X_j \in S}$.

Consider some iteration $i+1$ of OMP.
We will show that at least one of the $X'_i$ is a good candidate in
this iteration. 
Let $\ell$ maximize $\Rt{Z}{X'_\ell}$, i.e., 
$\ell \in \argmax_{(j:X'_j \in S'_i)}\Rt{Z}{X'_j}$.
By Lemma $3.7$,
\[
\Var{X'_\ell} 
\; \geq \; \eigmin{\COVM[S_G^{i} \cup \SET{X'_\ell}]} 
\; \geq \; \eigmink{\COVM}{2k}.
\] 
The OMP algorithm chooses a variable $X_m$ to add which maximizes
$\Abs{\Cov{\Residual{Z}{S_G^{i}}}{X_m}}$. Thus, $X_m$ maximizes
\[
\Cov{\Residual{Z}{S_G^{i}}}{X_m}^2 
\; = \; \Cov{Z}{\Residual{X_m}{S_G^{i}}}^2
\; = \; \Rt{Z}{\Residual{X_m}{S_G^{i}}} \cdot \Var{\Residual{X_m}{S_G^{i}}}.
\]
In particular, this implies
\[
\Rt{Z}{\Residual{X_m}{S_G^{i}}}
\; \geq \; \Rt{Z}{X'_\ell} \cdot \frac{\Var{X'_\ell}}{\Var{\Residual{X_m}{S_G^{i}}}}
\; \geq \; \Rt{Z}{X'_\ell} \cdot \frac{\eigmink{\COVM}{2k}}{\Var{\Residual{X_m}{S_G^{i}}}}
\; \geq \; \Rt{Z}{X'_\ell} \cdot \eigmink{\COVM}{2k},
\]
because $\Var{\Residual{X_m}{S_G^{i}}} \leq 1$.
As in the proof of Theorem \ref{thm:frapprox},
$\Rt{Z}{X'_\ell} \geq \frac{\superm{\OmpSet}{k}}{k} \cdot \Rt{Z}{S'_i}$, so
$\Rt{Z}{\Residual{X_m}{S_G^{i}}}
\geq \Rt{Z}{S'_i} \cdot \frac{\eigmink{\COVM}{2k} \cdot \superm{\OmpSet}{k}}{k}$.
With the same definition of $A(i)$ as in the previous proof,
we get that
$A(i+1) \geq \frac{\eigmink{\COVM}{2k}\superm{\OmpSet}{k}}{k}
(P - \sum_{j=1}^{i} A(j))$.
An inductive proof now shows that
\[
\Rt{Z}{S_G} 
\; = \; \sum_{i=1}^k A(i)
\; \geq \; (1-e^{-\eigmink{\COVM}{2k} \cdot \superm{\OmpSet}{k}})
\cdot \Rt{Z}{\OPTSET{k}}.
\]
The proofs of the other two inequalities follow the same pattern as the
proof for Forward Regression.
\end{proof}

\subsection{Oblivious Algorithm}
As a baseline, we also consider a greedy algorithm which 
completely ignores \COVM and simply selects the $k$ variables 
individually most correlated with $Z$.

\begin{definition}[Oblivious]
\label{def:oblivious}
The \todef{Oblivious} algorithm for subset selection is as follows:
Select the $k$ variables $X_i$ with the largest \CovV{i} values.
\end{definition}

Lemma \ref{lem:r2bounds} immediately implies
a simple bound for the Oblivious algorithm:

\begin{theorem}
The set $\OblSet$ selected by the Oblivious algorithm has the following
approximation guarantees:
\[
\Rt{Z}{\OblSet} 
\; \geq \; \frac{\superm{\emptyset}{k}}{\eigmaxk{\COVM}{k}} \cdot \OPT
\; \geq \; \frac{\eigmink{\COVM}{k}}{\eigmaxk{\COVM}{k}} \cdot \OPT.
\]
\end{theorem}

\begin{proof}
Let $S$ be the set chosen by the Oblivious algorithm, and
\OPTSET{k} the optimum set of $k$ variables.
By definition of the Oblivious algorithm,
$\sum_{i \in S} \Rt{Z}{X_i} \geq \sum_{i \in \OPTSET{k}}
\Rt{Z}{X_i}$,
so using Lemma \ref{lem:r2bounds},
we obtain that 
\[
\Rt{Z}{S} \geq \frac{\sum_{i\in S} \Rt{Z}{X_i}}{\eigmaxk{\COVM}{k}}
\; \geq \; \frac{\sum_{i \in \OPTSET{k}} \Rt{Z}{X_i}}{\eigmaxk{\COVM}{k}}
\; \geq \; \frac{\superm{\emptyset}{k}}{\eigmaxk{\COVM}{k}} \Rt{Z}{\OPTSET{k}}.
\]

The second inequality of the theorem follows directly from
Lemma \ref{lem:supermod}.
\end{proof}

\section{Dictionary Selection Bounds}\label{sec:dictionaryselection}

To demonstrate the wider applicability of the approximate
submodularity framework, we next obtain a tighter analysis for two
greedy algorithms for the dictionary selection problem, introduced
in \cite{krause}.

\subsection{The Algorithm \SDSMA}

The \SDSMA algorithm generalizes the Oblivious greedy 
algorithm to the problem of dictionary selection. 
It replaces the 
$\Rt{Z_j}{S}$ term in Definition \ref{def:dictionary-selection} 
with its modular approximation $\fma{Z_j}{S}=\sum_{i \in S}\Rt{Z_j}{X_i}$.
Thus, it greedily tries to maximize the function
$\FMA{D} = \sum_{j=1}^s \max_{S \subset D,\SetCard{S}=k}\fma{Z_j}{S}$,
over sets $D$ of size at most $d$; the inner maximum can be computed
efficiently using the Oblivious algorithm.

\begin{definition}[\SDSMA]
\label{def:sdsma}
The \todef{\SDSMA} algorithm for dictionary selection selects a 
dictionary $D$ of size $d$ iteratively 
as follows:
\begin{algorithmic}[1]
\STATE Initialize $D_0 = \emptyset$. 
\FOR{each iteration $i+1$}
\STATE Let $X_m$ be a variable maximizing 
\FMA{D \cup \SET{X_m}},
and set $S_{i+1} = S_i \cup \SET{X_m}$.
\ENDFOR
\STATE Output $D_d$.
\end{algorithmic}
\end{definition}

Using Lemma \ref{lem:r2bounds}, we can obtain the following 
multiplicative approximation guarantee for \SDSMA:

\begin{theorem} \label{thm:sdsmaapprox}
Let \DMa be the dictionary selected by the \SDSMA algorithm, and
\DOpt the optimum dictionary of size $\SetCard{D} \leq d$, 
with respect to the objective $F(D)$ from 
Definition \ref{def:dictionary-selection}. Then,
\[
F(\DMa) 
\; \geq \; \frac{\superm{\emptyset}{k}}{\eigmaxk{\COVM}{k}}(1 -\frac{1}{e}) \cdot F(\DOpt)
\; \geq \; \frac{\eigmink{\COVM}{k}}{\eigmaxk{\COVM}{k}}(1 -\frac{1}{e}) \cdot F(\DOpt).
\]
\end{theorem}

\begin{proof}
\Omit{Let $D_g$ be the dictionary selected by the \SDSMA algorithm.}
Let $\hat{D}$ be a dictionary of size $d$ 
maximizing \FMA{D}.
Because $\fma{Z_j}{S}$ is monotone and modular in $S$, 
$\FMAf$ is a monotone, submodular function. 
Hence, using the submodularity results of 
Nemhauser et al.~\cite{nemhauser:wolsey:fisher} and the optimality of 
$\hat{D}$ for $\FMAf$, 
\[ 
\FMA{\DMa} 
\; \geq \; \FMA{\hat{D}}(1-\frac{1}{e}) 
\; \geq \; \FMA{\DOpt}(1-\frac{1}{e}).
\]
Now, by applying Lemma \ref{lem:r2bounds} for each $Z_j$, it is easy to 
show that $\FMA{\DOpt} \geq \superm{\emptyset}{k} \cdot \FD{\DOpt}$, and
similarly $\FMA{\DMa} \leq \eigmaxk{\COVM}{k} \cdot \FD{\DMa}$. 
Thus we get 
$\FD{\DMa} \geq \frac{\superm{\emptyset}{k}}{\eigmaxk{\COVM}{k}}(1 -\frac{1}{e}) \FD{\DOpt}$.

The second part now follows from Lemma \ref{lem:supermod}.
\end{proof}
Note that these bounds significantly improve the previous additive
approximation guarantee obtained in \cite{krause}:
$\FD{\DMa} \geq (1-\frac{1}{e})\FD{\DOpt} - (2 - \frac{1}{e})k\cdot \mu(\COVM)$. 
In particular, when $\mu(\COVM) > \Theta(1/k)$, 
i.e., even just one pair of
variables has moderate correlation, the approximation
guarantee of Krause and Cevher becomes trivial.

\subsection{The Algorithm \SDSOMP}

We also obtain a multiplicative approximation guarantee for the greedy 
\SDSOMP algorithm, introduced by Krause and Cevher for
dictionary selection. Our bounds for \SDSOMP are much stronger than the 
additive bounds obtained by Krause and Cevher.
However, for both our results and theirs, the performance
guarantees for \SDSOMP are much weaker than those for \SDSMA.

The \SDSOMP algorithm generalizes the Orthogonal Matching Pursuit
algorithm for subset selection to the problem of dictionary
selection. In each iteration, it adds a new element to the currently
selected dictionary by using Orthogonal Matching Pursuit
to approximate the estimation of $\max_{\SetCard{S}=k}\Rt{Z_j}{S}$.

\begin{definition}[\SDSOMP]
\label{def:sdsomp}
The \todef{\SDSOMP} algorithm for dictionary selection selects a 
dictionary $D$ of size $d$ iteratively 
as follows:
\begin{algorithmic}[1]
\STATE Initialize $D_0 = \emptyset$. 
\FOR{each iteration $i+1$}
\STATE Let $X_m$ be a variable maximizing 
$\sum_{j=1}^s \Rt{Z_j}{S_{\text{OMP}}(D_i \cup \SET{X_m},Z_j,k)}$
where $S_{\text{OMP}}(D,Z,k)$ denotes the set selected by
Orthogonal Matching Pursuit for predicting $Z$ using $k$ variables
from $D$.
\STATE Set $S_{i+1} = S_i \cup \SET{X_m}$.
\ENDFOR
\STATE Output $D_d$.
\end{algorithmic}
\end{definition}

We now show how to obtain a multiplicative approximation guarantee
for \SDSOMP. The following definitions are key to our analysis; the
first two are from Definition \ref{def:dictionary-selection} and 
Theorem \ref{thm:sdsmaapprox}.
\begin{align*}
\FD{D} & =  \sum_{j=1}^s \max_{S \subset D,\SetCard{S}=k}\Rt{Z_j}{S},\\
\FMA{D} & =  \sum_{j=1}^s \max_{S \subset D,\SetCard{S}=k}\fma{Z_j}{S},\\
\FOMP{D} & =  \sum_{j=1}^s \Rt{Z_j}{S_{\text{OMP}}(D,Z_j,k)}.
\end{align*}

We first prove
the following lemma about approximating the function $\FMA{D}$ by
$\FOMP{D}$:

\begin{lemma}
\label{lem:approxsds}
For any set $D$, we have that
\[ \begin{array}{lclclcl}
\frac{(1 - e^{-\eigmink{\COVM}{2k}^2})}{\eigmaxk{\COVM}{k}}\cdot\FMA{D}
& \leq & \FOMP{D} & \leq &\frac{\FMA{D}}{\superm{\emptyset}{k}}.
\end{array} \]
\end{lemma}

\begin{proof}
Using Theorem \ref{thm:ompapprox} and  Lemma \ref{lem:r2bounds} and
summing up over all the $Z_j$ terms, we obtain that
\[ \begin{array}{lclcl}
\FOMP{D}
& \geq & (1 - e^{-\eigmink{\COVM}{2k}^2}) \cdot \FD{D}
& \geq & (1 - e^{-\eigmink{\COVM}{2k}^2})
\frac{\FMA{D}}{\eigmaxk{\COVM}{k}}.
\end{array} \]

Similarly, using Lemma \ref{lem:r2bounds} and the fact that
$\max_{S \subset D,\SetCard{S}=k}\Rt{Z_j}{S} \geq
\Rt{Z_j}{S_{OMP}(D,Z_j,k)}$,
we have
\[ \begin{array}{lclclcl}
\FMA{D} & \geq &
\superm{\emptyset}{k} \cdot \FD{D} & \geq & \superm{\emptyset}{k} \cdot \FOMP{D}.
\end{array} \]
\end{proof}

Using the above lemma, we now prove the following bound for \SDSOMP:

\begin{theorem} \label{thm:sdsompapprox}
Let \DOmp be the dictionary selected by the \SDSOMP algorithm,
and \DOpt the optimum dictionary of size $\SetCard{D} \leq d$,
with respect to the objective $F(D)$ from
Definition \ref{def:dictionary-selection}. Then,
\[ 
\FD{\DOmp} 
\; \geq \; \FD{\DOpt} \cdot
\frac{\superm{\emptyset}{k}}{\eigmaxk{\COVM}{k}} \cdot
\frac{(1 - e^{-(p \cdot \superm{\emptyset}{k})})}
{d - d\cdot p \cdot \superm{\emptyset}{k} + 1} 
\; \geq \;
\FD{\DOpt} \cdot
\frac{\eigmink{\COVM}{k}}{\eigmaxk{\COVM}{k}} \cdot
\frac{(1 - e^{-(p \cdot \superm{\emptyset}{k})})}
{d - d\cdot p \cdot \superm{\emptyset}{k} + 1},
\]
where 
$p = \frac{1}{\eigmaxk{\COVM}{k}} \cdot (1 - e^{-\eigmink{\COVM}{2k}^2})$.
\end{theorem}

\begin{proof}
Let $\hat{D}$ be the dictionary of
size $d$ that maximizes \FMA{D}.
We first prove that \FMA{\DOmp} is a good approximation to
\FMA{\hat{D}}.

Let $\GrSet{i}$ be the variables chosen by \SDSOMP
after $i$ iterations. Define $S_i = \hat{D} \setminus \GrSet{i}$.
By monotonicity of $\FMAf$, we have that
$\FMA{S_i \cup \GrSet{i}} \geq  \FMA{\hat{D}}$.

Let $\hat{X} \in S_i$ be the variable maximizing
$\FMA{\GrSet{i} \cup \SET{\hat{X}}}$, and similarly
$\tilde{X} \in S_i$ be the variable maximizing
$\FOMP{\GrSet{i} \cup \SET{\tilde{X}}}$.

Since \FMAf is a submodular function, it is easy to show (using
an argument similar to the proof of Theorem \ref{thm:frapprox}) that
$\FMA{\GrSet{i} \cup \SET{\hat{X}}} - \FMA{\GrSet{i}} \geq
\frac{\FMA{\hat{D}} - \FMA{\GrSet{i}}}{d}$.

Now, using Lemma \ref{lem:approxsds} above, and
the optimality of $\tilde{X}$ for $\FOMP{\GrSet{i} \cup \SET{\tilde{X}}}$,
we obtain that
\[ \begin{array}{lclclclclcl}
\frac{1}{\superm{\emptyset}{k}} \cdot
\FMA{\GrSet{i} \cup \SET{\tilde{X}}}
& \geq & \FOMP{\GrSet{i} \cup \SET{\tilde{X}}} & \geq &
\FOMP{\GrSet{i} \cup \SET{\hat{X}}} & \geq &
p \cdot \FMA{\GrSet{i} \cup \SET{\hat{X}}}.
\end{array} \]

Thus, $\FMA{\GrSet{i} \cup \SET{\tilde{X}}} \geq
p \cdot \superm{\emptyset}{k} \cdot \FMA{\GrSet{i} \cup \SET{\hat{X}}}$, or
\begin{align*}
\FMA{\GrSet{i} \cup \SET{\tilde{X}}} - \FMA{\GrSet{i}} 
& \geq 
p \cdot \superm{\emptyset}{k} \cdot
(\FMA{\GrSet{i} \cup \SET{\hat{X}}} - \FMA{\GrSet{i}})
 - (1 - p \cdot \superm{\emptyset}{k}) \FMA{\GrSet{i}}.
\end{align*}

Define $A(i) = \FMA{\GrSet{i}} - \FMA{\GrSet{i-1}}$ to be the gain, with
respect to \FMAf,
obtained from the variable chosen by \SDSOMP in iteration $i$.
Then $\FMA{\DOmp} = \sum_{i=1}^d A(i)$.
From the preceding paragraphs, we obtain
\begin{align*}
A(i+1) & \geq  \frac{p \cdot \superm{\emptyset}{k}}{d} \cdot
(\FMA{\hat{D}} -
(1 + \frac{d}{p \cdot \superm{\emptyset}{k}} - d)\sum_{j=1}^{i} A(j)).
\end{align*}

Since the above inequality holds for each iteration $i=1,2,\ldots, d$, a
simple inductive proof shows that
\begin{align*}
\FMA{\hat{D}} - \sum_{i=1}^{d} A(i) 
& \leq
\FMA{\hat{D}} \cdot (1 - \frac{p \superm{\emptyset}{k}}{d})^d +
(d - d p \superm{\emptyset}{k}) \cdot \sum_{i=1}^{d} A(i).
\end{align*}

Rearranging the terms and simplifying, we get that
\[ \begin{array}{lclclclclcl}
\FMA{\DOmp} & = & \sum_{i=1}^{d} A(i) & \geq & \FMA{\hat{D}} \cdot
\frac{(1 - e^{-(p \cdot \superm{\emptyset}{k})})}
{d - d p \superm{\emptyset}{k} + 1}
& \geq &
\FMA{\DOpt} \cdot
\frac{(1 - e^{-(p \cdot \superm{\emptyset}{k})})}
{d - d p  \superm{\emptyset}{k} + 1},
\end{array} \]
where the last inequality is due to the optimality of $\hat{D}$ for
$\FMAf$.

Now, using Lemma \ref{lem:r2bounds} for each $Z_j$ term, it can be easily seen
that $\FMA{\DOpt} \geq \superm{\emptyset}{k} \cdot \FD{\DOpt}$.
Similarly, using Lemma $3.3$ on the set $\DOmp$, we have
$\FD{\DOmp} \geq \frac{1}{\eigmaxk{\COVM}{k}} \cdot \FMA{\DOmp}$.

Using the above inequalities, we therefore get the desired bound
\begin{align*}
\FD{\DOmp} 
& \geq  \FD{\DOpt} \cdot
\frac{\superm{\emptyset}{k}}{\eigmaxk{\COVM}{k}} \cdot
\frac{(1 - e^{-(p \cdot \superm{\emptyset}{k})})}
{d - d\cdot p \cdot \superm{\emptyset}{k} + 1}.
\end{align*}

The second inequality of the Theorem now follows directly from
Lemma \ref{lem:supermod}.
\end{proof}


\section{Experiments} \label{sec:experiments}
In this section, we evaluate Forward Regression (FR) and OMP empirically,
on two real-world and one synthetic data set. We compare the
two algorithms against an optimal solution (OPT), computed using exhaustive
search, the Oblivious greedy algorithm (OBL), and the
L1-regularization/Lasso (L1) algorithm 
(in the implementation of Koh et al.~\cite{boydsw}).
Beyond the algorithms' performance, we also compute the various
spectral parameters from which we can 
derive lower bounds. Specifically, these are
\begin{enumerate}
\item the submodularity ratio: $\superm{\FrSet}{k}$, where \FrSet is the subset
selected by forward regression.
\item the smallest sparse eigenvalues $\eigmink{\COVM}{k}$ and
$\eigmink{\COVM}{2k}$. (In
some cases, computing $\eigmink{\COVM}{2k}$ was not
computationally feasible due to the problem size.)
\item the sparse inverse condition number
$\CondNumk{\COVM}{k}^{-1}$.
As mentioned earlier, the sparse inverse condition number 
$\CondNumk{\COVM}{k}$ is strongly related to the Restricted Isometry 
Property in \cite{candes}.
\item the smallest eigenvalue $\eigmin{\COVM} = \eigmink{\COVM}{n}$
of the entire covariance matrix.
\end{enumerate}

The aim of our experiments is twofold: 
First, we wish to evaluate which among the submodular and
spectral parameters are good predictors of the performance of greedy
algorithms in practice.
Second, we wish to highlight how the theoretical bounds for subset
selection algorithms reflect on their actual performance.
Our analytical results predict
that Forward Regression should outperform OMP, which in turn
outperforms Oblivious. For Lasso, it is not
known whether strong multiplicative bounds, like the ones we proved for 
Forward Regression or OMP, can be obtained. 

\subsection{Data Sets}
Because several of the spectral parameters (as well as the optimum
solution) are \NP-hard to compute, we restrict our experiments to data
sets with $n \leq 30$ features, from which $k \leq 8$ are to be
selected. We stress that the greedy algorithms themselves are
very efficient, and the restriction on data set sizes is only
intended to allow for an adequate evaluation of the results.

Each data set contains $m > n$ samples, from which we compute the 
empirical covariance matrix (analogous to
the Gram matrix in sparse approximation) between all
observation variables and the predictor variable; we then normalize
it to obtain \COVM and \COVV.
We evaluate the performance of all algorithms in terms of their $R^2$
fit; thus, we implicitly treat \COVM and \COVV as the ground truth,
and also do not separate the data sets into training and test cases.

Our data sets are the \emph{Boston Housing Data}, a data set of
\emph{World Bank Development Indicators}, and a synthetic data set
generated from a distribution similar to the one used by Zhang~\cite{zhangfb}.
The \emph{Boston Housing Data} (available from the UCI Machine
Learning Repository) is a small data set frequently used to evaluate
ML algorithms. It comprises $n=15$ features (such as crime rate, 
 property tax rates, etc.) and $m=516$ observations.
Our goal is to predict housing prices from these features.
The \emph{World Bank Data} (available from \texttt{http://databank.worldbank.org})
contains  an extensive list of socio-economic and health
indicators of development, for many countries and over several years.
We choose a subset of $n=29$ indicators for the years 2005 and 2006,
such that the values for all of the $m=65$ countries are known for
each indicator. (The data set does not contain all indicators for each
country.)
We choose to predict the average life expectancy for those countries.

To perform tests in a controlled fashion,
we also generate random instances from a known distribution
similar to \cite{zhangfb}: There are $n=29$ features, and $m=100$ data
points are generated from a joint Gaussian distribution with
moderately high correlations of $0.6$.
The target vector is obtained by generating coefficients uniformly from $0$
to $10$ along each dimension, and adding noise with variance $\sigma^2 = 0.1$.
Notice that the target vector is not truly sparse. 
The plots we show are the average $R^2$ values
for 20 independent runs of the experiment.

\subsection{Results}

\begin{figure}[ht]
\begin{minipage}[b]{0.45\linewidth}
\begin{center}
\epsfxsize=8cm
\epsffile{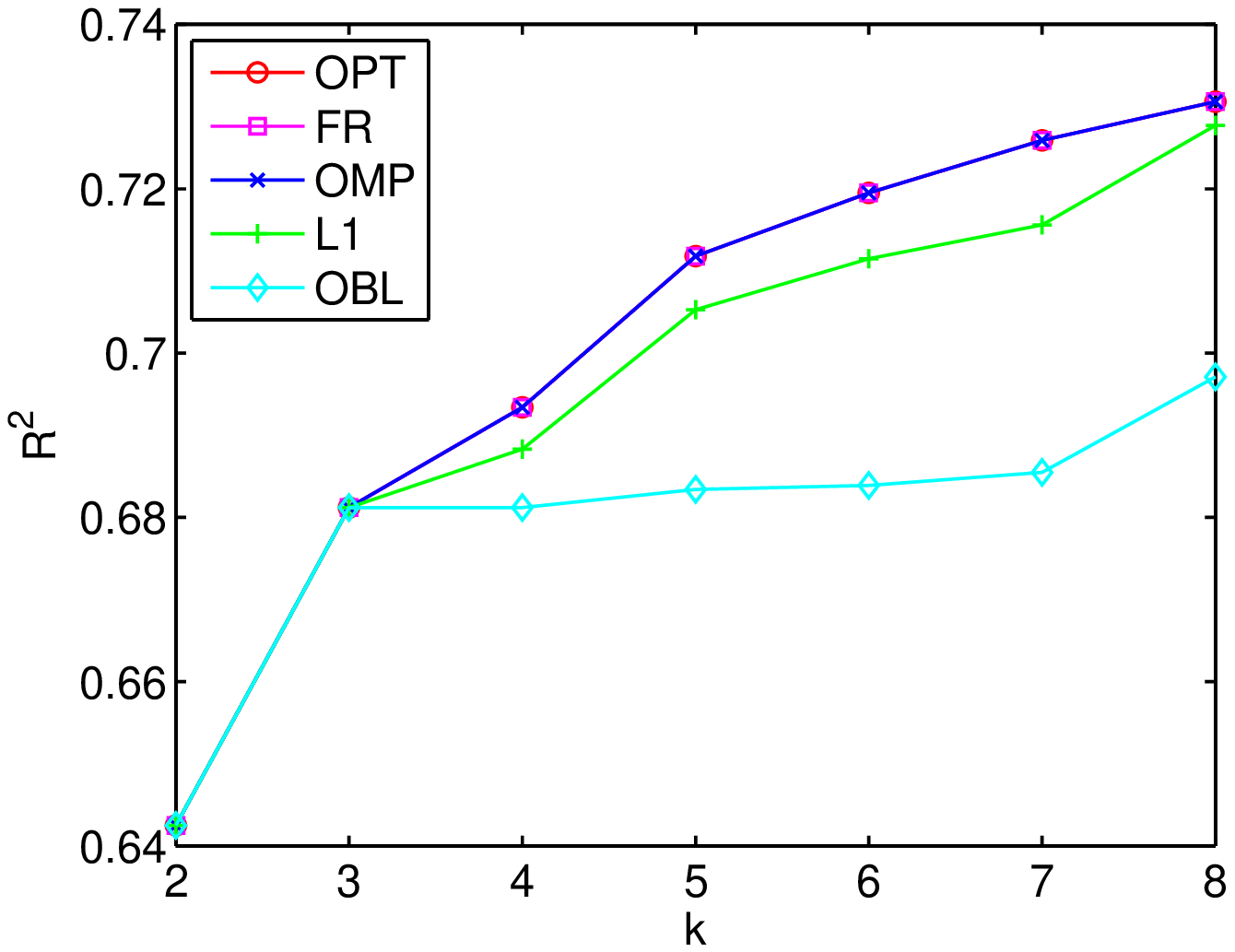}
\caption{Boston Housing $R^2$ \label{fig:bostonr}}
\end{center}
\end{minipage}
\hspace{0.5cm}
\begin{minipage}[b]{0.45\linewidth}
\begin{center}
\epsfxsize=8cm
\epsffile{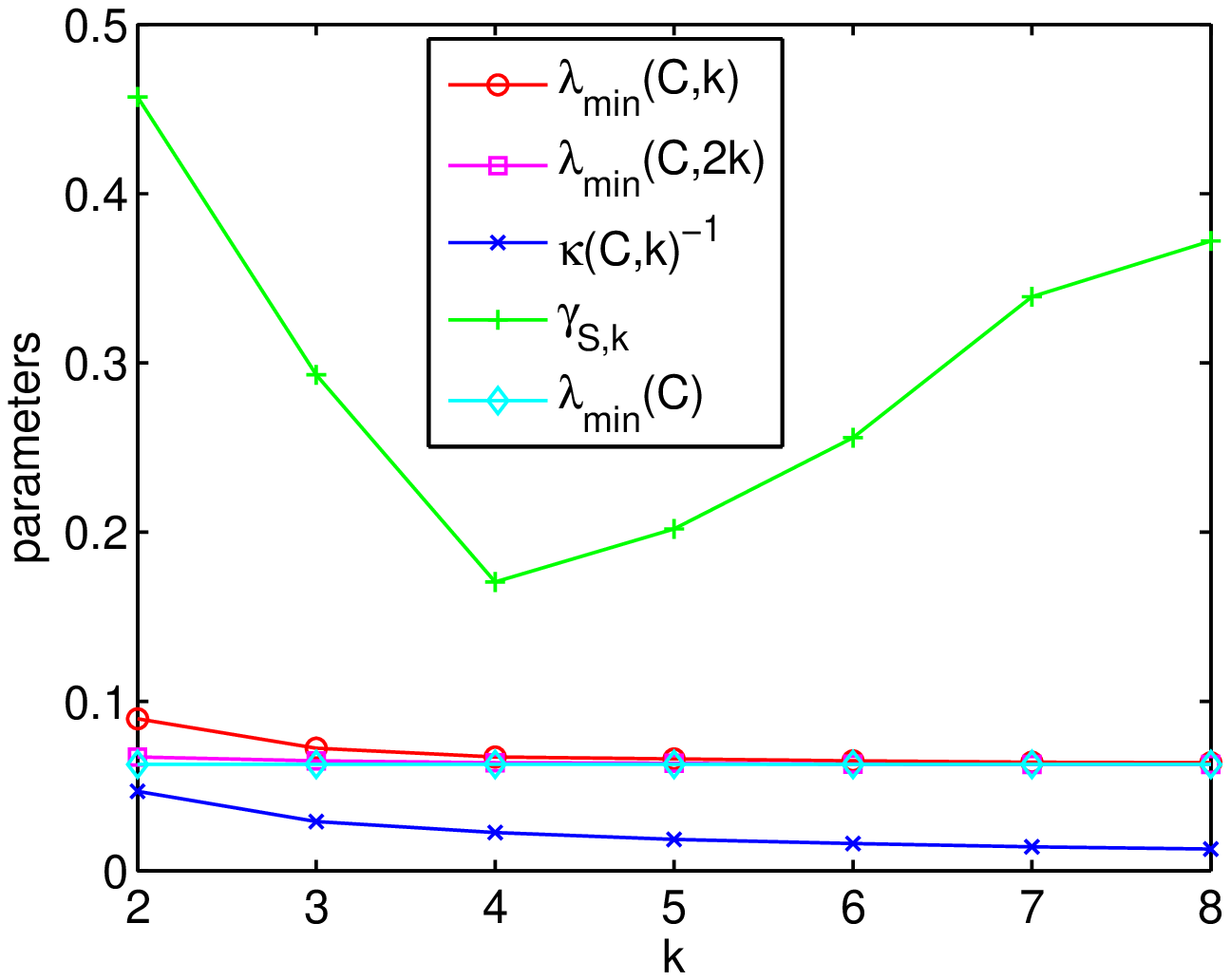}
\caption{Boston Housing parameters \label{fig:bostonl}}
\end{center}
\end{minipage}
\end{figure}

\begin{figure}[ht]
\begin{minipage}[b]{0.45\linewidth}
\begin{center}
\epsfxsize=8cm
\epsffile{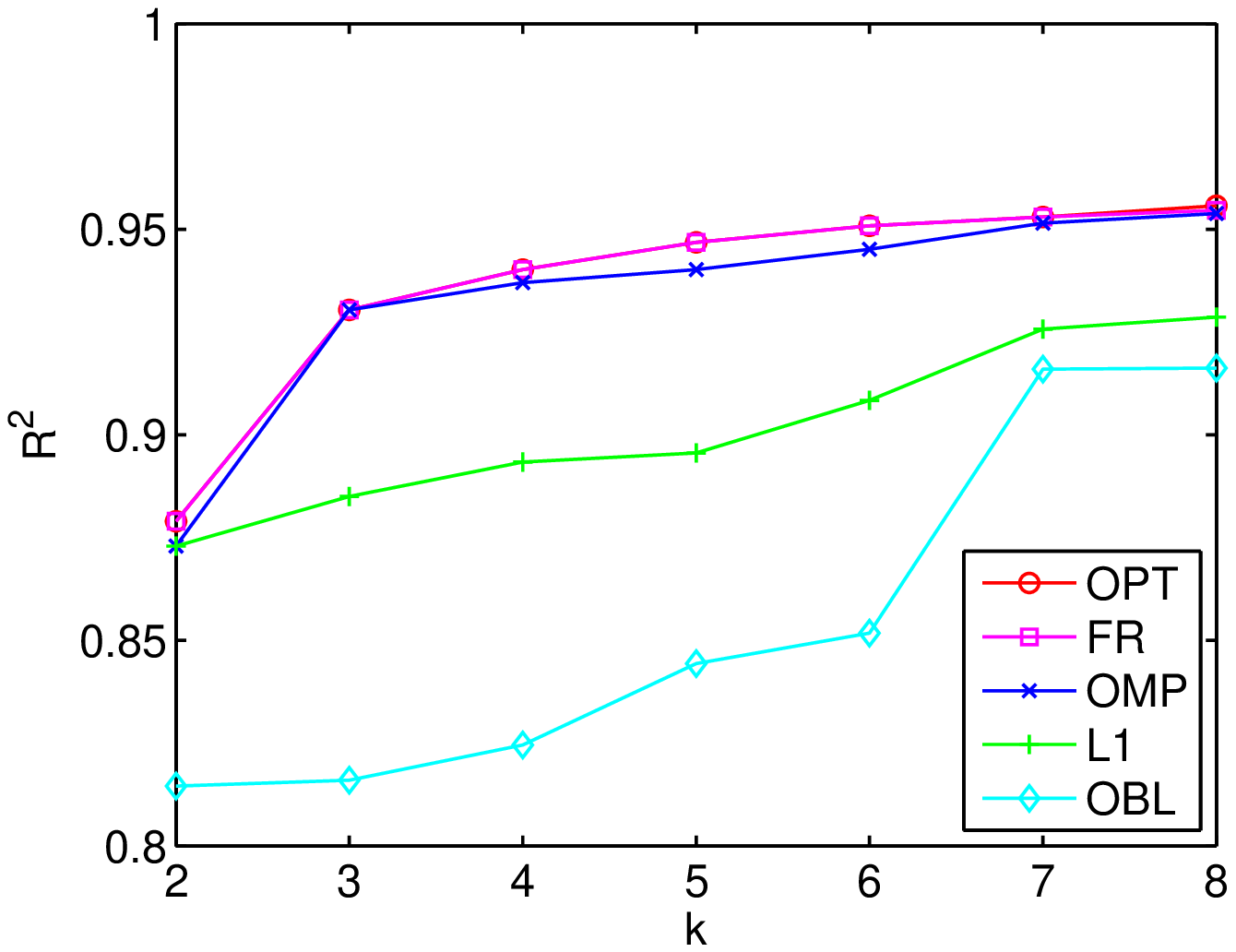}
\caption{World Bank $R^2$ \label{fig:worldbankr}}
\end{center}
\end{minipage}
\hspace{0.5cm}
\begin{minipage}[b]{0.45\linewidth}
\begin{center}
\epsfxsize=8cm
\epsffile{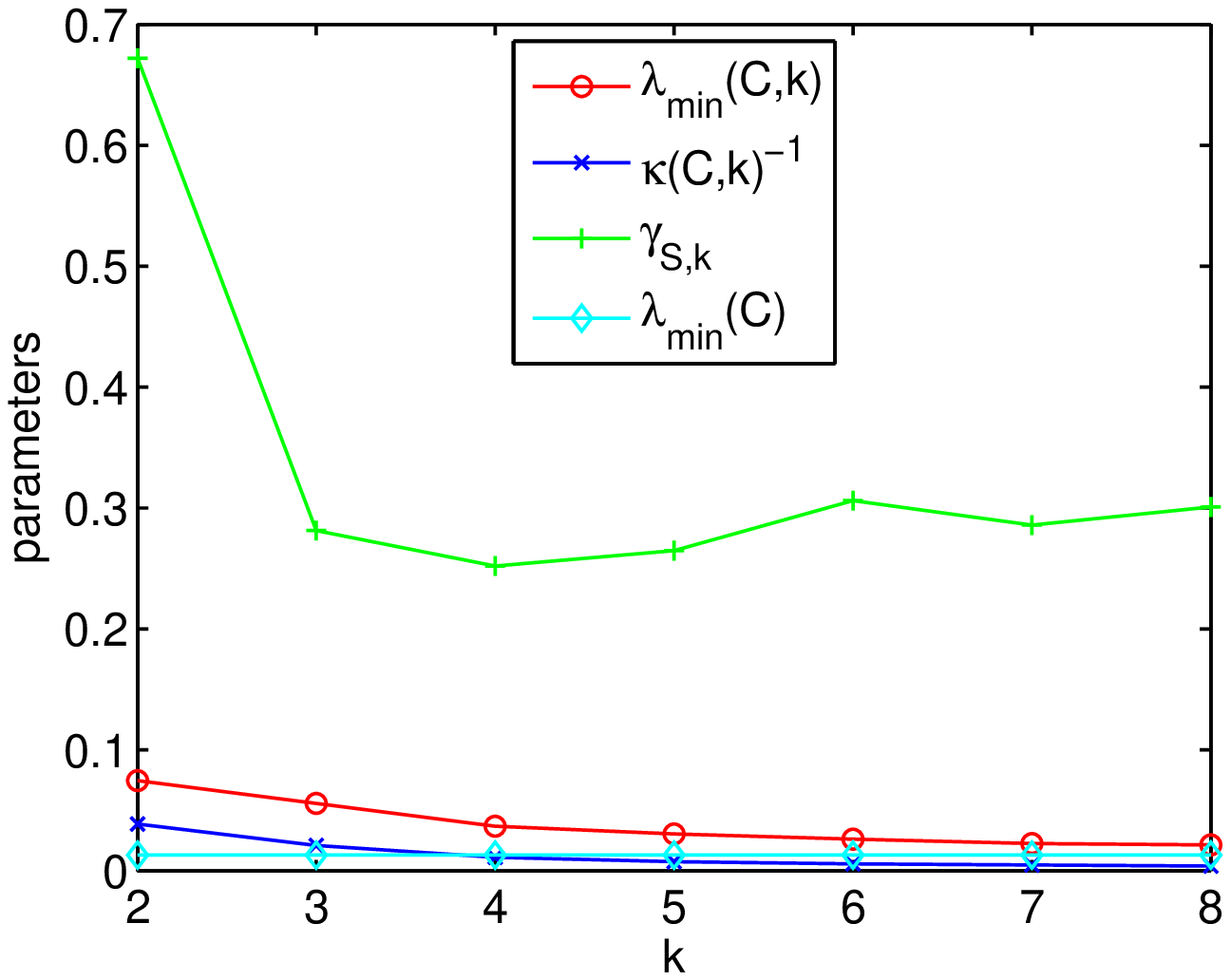}
\caption{World Bank parameters \label{fig:worldbankl}}
\end{center}
\end{minipage}
\end{figure}

\begin{figure}[ht]
\begin{minipage}[b]{0.45\linewidth}
\begin{center}
\epsfxsize=8cm
\epsffile{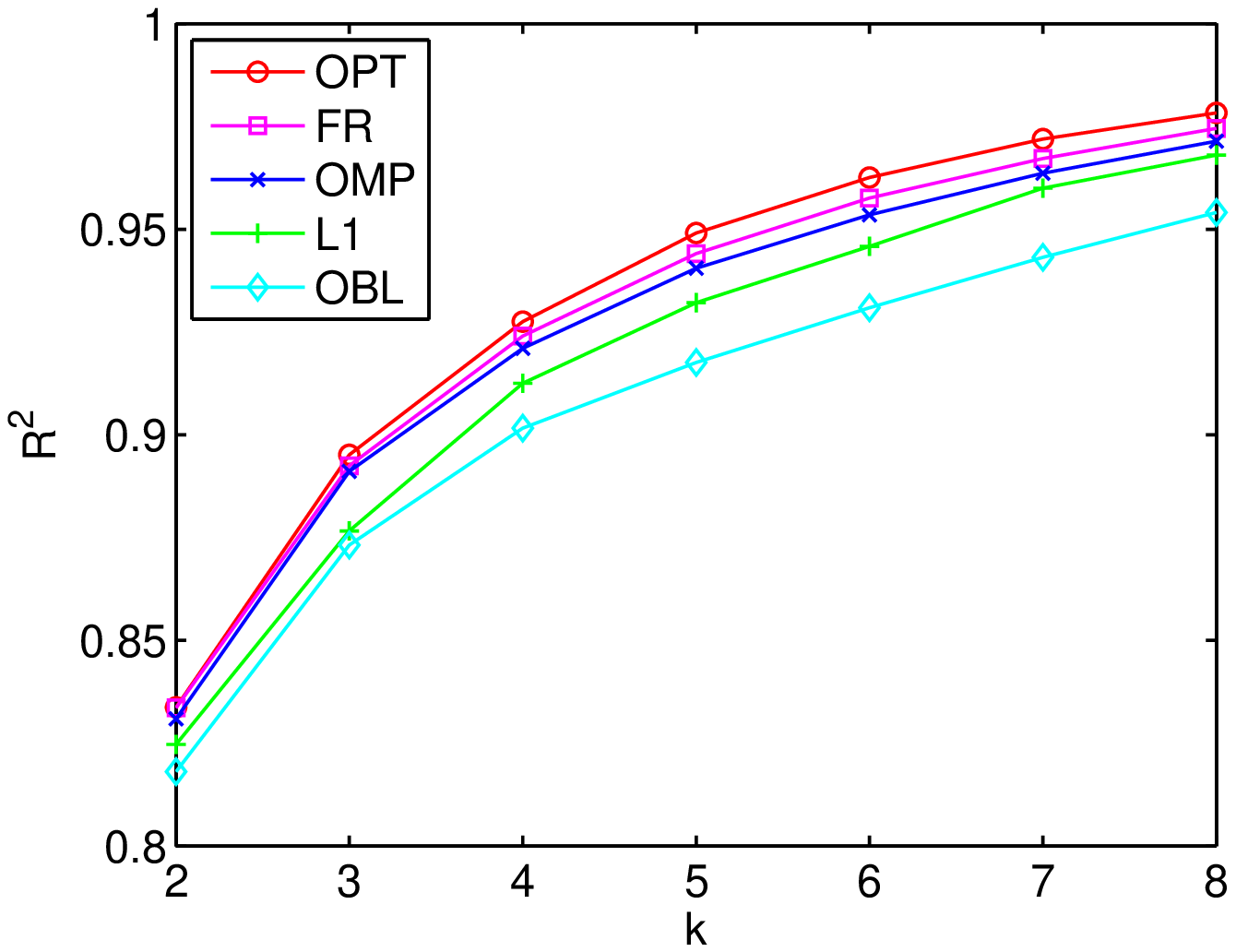}
\caption{Synthetic Data $R^2$ \label{fig:randomr}}
\end{center}
\end{minipage}
\hspace{0.5cm}
\begin{minipage}[b]{0.45\linewidth}
\begin{center}
\epsfxsize=8cm
\epsffile{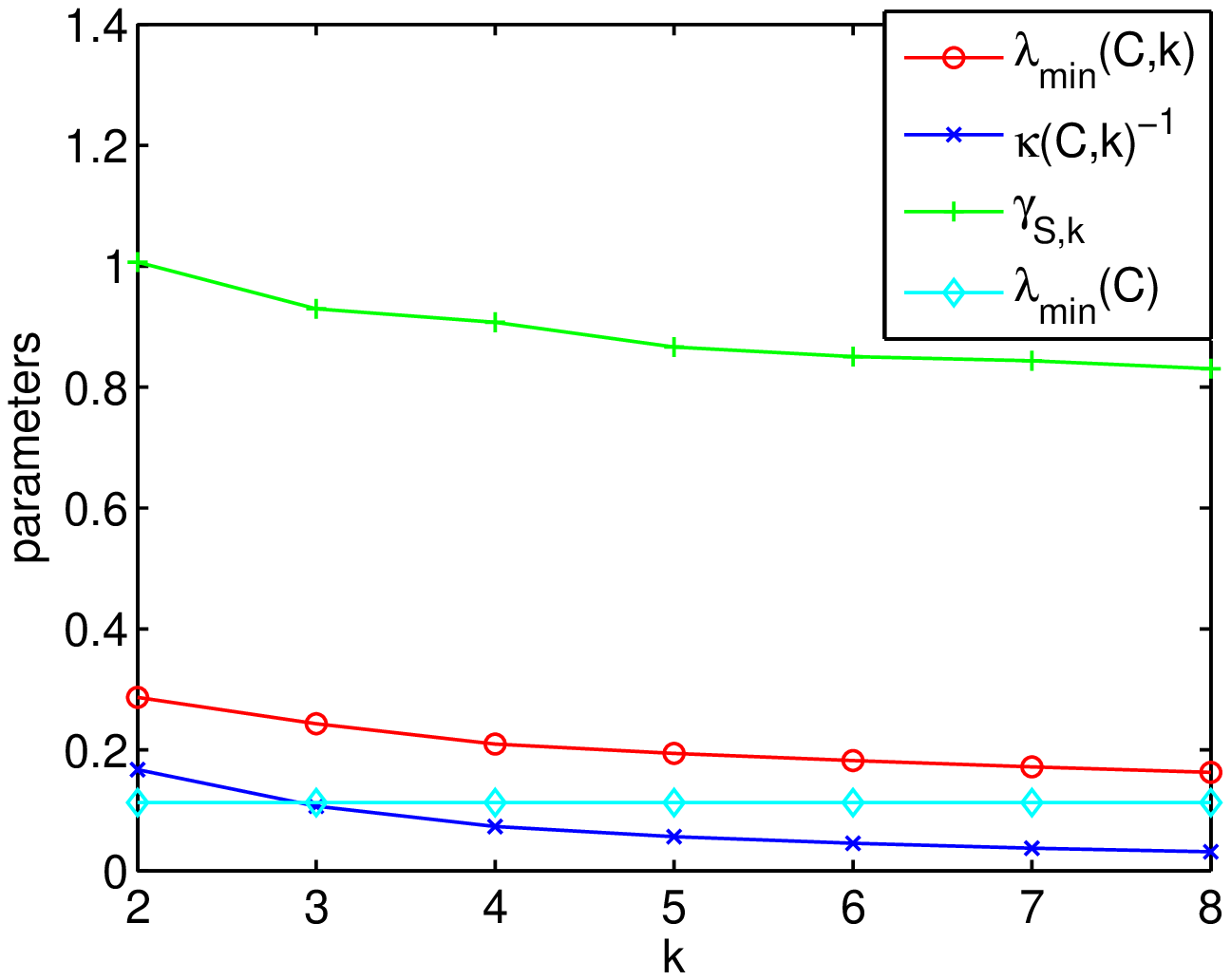}
\caption{Synthetic Data parameters \label{fig:randoml}}
\end{center}
\end{minipage}
\end{figure}

We run the different subset selection algorithms for values of $k$
from $2$ through $8$, and plot the $R^2$ values for the selected
sets.\remove{We notice that the $R^2$ value when including all of the
features is close to 1 in all data sets, meaning that nearly all of
the variance in the function to be predicted can be explained by the
features.}
Figures
\ref{fig:bostonr}, \ref{fig:worldbankr} and \ref{fig:randomr} show the
results for the three data sets.
The main insight is that on all data sets, Forward Regression performs
optimally or near-optimally, and OMP is
only slightly worse. 
Lasso performs somewhat worse on all data sets,
and, not surprisingly, the baseline Oblivious algorithm 
performs even worse.
The order of performance of the greedy algorithms match the order 
of the strength of the theoretical bounds we derived for them.

On the World Bank data (Figure \ref{fig:worldbankr}), all algorithms
perform quite well with just 2--3 features already. The main reason is
that adolescent birth rate is by itself highly predictive of life
expectancy, so the first feature selected by all algorithms already
contributes high $R^2$ value.

Figures \ref{fig:bostonl}, \ref{fig:worldbankl} and \ref{fig:randoml} show the different
spectral quantities for the data sets, for varying values of $k$.
Both of the real-world data sets are nearly singular, as evidenced by
the small \eigmin{\COVM} values. In fact, the near singularities
manifest themselves for small values of $k$ already; in particular,
since \eigmink{\COVM}{2} is already small, we observe that there are
pairs of highly correlated observations variables in the data sets.
Thus, the bounds on approximation we would obtain by considering 
merely \eigmink{\COVM}{k} or \eigmink{\COVM}{2k} would be quite weak.
Notice, however, that they are still quite  a bit stronger than the
inverse condition number $\CondNumk{\COVM}{k}^{-1}$: this bound --- 
which is closely related to the RIP property frequently at
the center of sparse approximation analysis --- takes on much smaller
values, and thus would be an even looser bound than the eigenvalues.

On the other hand, the submodularity ratios \superm{\FrSet}{k} 
for all the data sets are much larger than the other spectral quantities 
(almost $5$ times larger, on average, than the corresponding 
\eigmin{\COVM} values). Notice that unlike the other quantities, 
the submodularity ratios are not monotonically decreasing in $k$ --- 
this is due to the dependency of \superm{\FrSet}{k} on the set $\FrSet$, 
which is different for every $k$.

The discrepancy between the small values of the eigenvalues
and the good performance of all algorithms shows that bounds based
solely on eigenvalues can sometimes be loose. Significantly better bounds
are obtained from the submodularity ratio \superm{\FrSet}{k}, which takes
on values above 0.2, and significantly larger in many cases. While not
entirely sufficient to explain the performance of the greedy
algorithms, it shows that the near-singularities of \COVM do not align
unfavorably with \COVV, and thus do not provide an opportunity for
strong supermodular behavior that adversely affects greedy algorithms.

The synthetic data set we generated is somewhat further from singular,
with $\eigmin{\COVM} \approx 0.11$. However, the same patterns
persist: the simple eigenvalue based bounds, while somewhat larger for
small $k$, still do not fully predict the performance of 
greedy algorithms,
whereas the submodularity ratio here is close to 1 for all values of
$k$. This shows that the near-singularities do not at all provide the
possibility of strongly supermodular benefits of sets of
variables. Indeed, the plot of $R^2$ values on the synthetic data is
concave, an indicator of submodular behavior of the 
function.

The above observations suggest that bounds based on the 
submodularity ratio are better predictors of the performance of greedy 
algorithms, followed by bounds based on the sparse
eigenvalues, and finally those based on the condition number or RIP property.

\subsection{Narrowing the gap between theory and practice}

Our theoretical bounds, though much stronger than 
previous results, still do not fully predict the observed 
near-optimal performance of Forward Regression and OMP on the
real-world datasets. 
In particular, for Forward Regression, even though 
the submodularity ratio is less than $0.4$ for most cases,
implying a theoretical guarantee of roughly $1 - e^{-0.4} \approx 33\%$, 
the algorithm still achieves near-optimal performance. 
While gaps between worst-case bounds and practical performance
are commonplace in algorithmic analysis, they also
suggest that there is scope for further improving the analysis,
by looking at more fine-grained parameters. 

Indeed, a slightly more careful analysis of the proof of 
Theorem \ref{thm:frapprox} and our definition of the submodularity ratio
reveals that we do not really need to calculate the submodularity ratio 
over all sets $S$ of size $k$ while analyzing the greedy steps of Forward Regression.
We can ignore sets $S$ whose
submodularity ratio is low, but whose marginal contribution to the
current $R^2$ is only a small fraction (say, at most $\epsilon$). 
This is because the proof of Theorem \ref{thm:frapprox} shows that for
each iteration $i+1$, we only need to consider the submodularity ratio
for the set $S_i = \OPTSET{k} \setminus \GrSet{i}$, where \GrSet{i} is
the set selected by the greedy algorithm after $i$ iterations, 
and \OPTSET{k} is the optimal $k$-subset. 
Thus, if $\Rt{Z}{S_i \cup \GrSet{i}} \leq (1+ \epsilon) \cdot 
\Rt{Z}{\GrSet{i}}$,
then the currently selected set must already be 
within a factor $\frac{1}{1+\epsilon}$ of optimal.

By carefully pruning such sets (using $\epsilon = 0.2$) while 
calculating the submodularity ratio, we see that the resulting values
of $\superm{\FrSet}{k}$ are much higher (more than $0.8$), 
thus significantly reducing the gap between the theoretical bounds 
and experimental results. Table \ref{table:newsuperm} 
shows the values of $\superm{\FrSet}{k}$ obtained using this method.

The results suggest an interesting direction for future work: 
namely, to characterize for which sets the
submodular behavior of $R^2$ really matters.

\begin{table}[ht]
\centering
\begin{tabular}{l r r r r r r r}
\hline\hline
Data Set & $k=2$ & $k=3$ & $k=4$ & $k=5$ & $k=6$ & $k=7$ & $k=8$\\
\hline
Boston & 0.9 & 0.91 & 1.02 & 1.21 & 1.36 & 1.54 & 1.74\\
World Bank & 0.8 & 0.81 & 0.81 & 0.81 & 0.94 & 1.19 & 1.40 \\
\hline
\end{tabular}
\caption{Improved estimates for submodularity ratio
\label{table:newsuperm}}
\end{table}

\section{Discussion and Concluding Remarks}
In this paper, we analyze greedy algorithms using the notion of 
submodularity ratio, which captures how close to submodular an
objective function (in our case the $R^2$ measure of statistical fit) is. 
Using submodular analysis, coupled with spectral 
techniques, we prove the strongest known approximation guarantees 
for commonly used greedy algorithms for subset selection
and dictionary selection. Our bounds help explain why greedy algorithms
perform well in practice even in the presence of strongly correlated 
data, and are substantiated by experiments on real-world and synthetic
datasets. The experiments show that the submodularity ratio is a much
stronger predictor of the performance of greedy algorithms than 
previously used spectral parameters. 
  We believe that our techniques for analyzing greedy algorithms
using a notion of ``approximate submodularity'' are not specific to
subset selection and dictionary selection, and could also be used to 
analyze other problems in compressed sensing and sparse recovery. 

\bibliographystyle{plain}
\bibliography{icml}

\begin{thebibliography}{10}

\bibitem{candes}
E.~J. Cand\`{e}s, J.~Romberg, and T.~Tao.
\newblock Stable signal recovery from incomplete and inaccurate measurements.
\newblock {\em {Communications on Pure and Applied Mathematics}},
  59:1207--1223, 2005.

\bibitem{das:kempe}
A.~Das and D.~Kempe.
\newblock Algorithms for subset selection in linear regression.
\newblock In {\em ACM Symposium on Theory of Computing}, 2008.

\bibitem{diekhoff}
G.~Diekhoff.
\newblock {\em Statistics for the Social and Behavioral Sciences}.
\newblock Wm. C. Brown Publishers, 2002.

\bibitem{donoho}
D.~Donoho.
\newblock For most large underdetermined systems of linear equations, the
  minimal 11-norm near-solution approximates the sparsest near-solution.
\newblock {\em {Communications on Pure and Applied Mathematics}},
  59:1207--1223, 2005.

\bibitem{muthu}
A.~Gilbert, S.~Muthukrishnan, and M.~Strauss.
\newblock Approximation of functions over redundant dictionaries using
  coherence.
\newblock In {\em {Proc. ACM-SIAM Symposium on Discrete Algorithms}}, 2003.

\bibitem{johnson}
R.~A. Johnson and D.~W. Wichern.
\newblock {\em Applied Multivariate Statistical Analysis}.
\newblock Prentice Hall, 2002.

\bibitem{boydsw}
K.~Koh, S.~Kim, and S.~Boyd.
\newblock {l1\_ls: Simple Matlab Solver for l1-regularized Least Squares
  Problems}, 2008.
\newblock http://www.stanford.edu/~boyd/l1\_ls.

\bibitem{krause}
A.~Krause and V.~Cevher.
\newblock Submodular dictionary selection for sparse representation.
\newblock In {\em {Proc. ICML}}, 2010.

\bibitem{lozano}
A.~C. Lozano, G.~Swirszcz, and N.~Abe.
\newblock Grouped orthogonal matching pursuit for variable selection and
  prediction.
\newblock In {\em {Proc. NIPS}}, 2009.

\bibitem{miller}
A.~Miller.
\newblock {\em Subset Selection in Regression}.
\newblock Chapman and Hall, second edition, 2002.

\bibitem{natarajan}
B.~Natarajan.
\newblock Sparse approximation solutions to linear systems.
\newblock {\em {SIAM Journal on Computing}}, 24:227--234, 1995.

\bibitem{nemhauser:wolsey:fisher}
G.~Nemhauser, L.~Wolsey, and M.~Fisher.
\newblock An analysis of the approximations for maximizing submodular set
  functions.
\newblock {\em Mathematical Programming}, 14:265--294, 1978.

\bibitem{tibshirani}
R.~Tibshirani.
\newblock Regression shrinkage and selection via the lasso.
\newblock {\em {Journal of Royal Statistical Society}}, 58:267--288, 1996.

\bibitem{tropp3}
J.~Tropp.
\newblock Greed is good: algorithmic results for sparse approximation.
\newblock {\em {IEEE Trans. Information Theory}}, 50:2231--2242, 2004.

\bibitem{tropp4}
J.~Tropp.
\newblock Just relax: Convex programming methods for identifying sparse
  signals.
\newblock {\em {IEEE Trans. Information Theory}}, 51:1030--1051, 2006.

\bibitem{tropp2}
J.~Tropp, A.~Gilbert, S.~Muthukrishnan, and M.~Strauss.
\newblock Improved sparse approximation over quasi-incoherent dictionaries.
\newblock In {\em {Proc. IEEE-ICIP}}, 2003.

\bibitem{zhangfb}
T.~Zhang.
\newblock Adaptive forward-backward greedy algorithm for sparse learning with
  linear models.
\newblock In {\em {Proc. NIPS}}, 2008.

\bibitem{zhangf}
T.~Zhang.
\newblock On the consistency of feature selection using greedy least squares
  regression.
\newblock {\em {Journal of Machine Learning Research}}, 10:555--568, 2009.

\bibitem{yu}
P.~Zhao and B.~Yu.
\newblock On model selection consistency of lasso.
\newblock {\em {Journal of Machine Learning Research}}, 7:2451--2457, 2006.

\bibitem{zhou}
S.~Zhou.
\newblock Thresholding procedures for high dimensional variable selection and
  statistical estimation.
\newblock In {\em {Proc. NIPS}}, 2009.

\end{thebibliography}


\begin{appendix}
\section{Estimating \eigmink{\COVM}{k}}
\label{app:eigptas}
Several of our approximation guarantees are phrased in terms of
\eigmink{\COVM}{k}.
Finding the exact value of $\eigmink{\COVM}{k}$ is \NP-hard in
general; here, we show how to estimate lower and upper bounds.
Let $\lambda_1 \leq \lambda_2 \leq \ldots \leq \lambda_n$ be the
eigenvalues of \COVM,
and $\Vector{e_1}, \Vector{e_2}, \ldots, \Vector{e_n}$
the corresponding eigenvectors.
A first simple bound can be obtained directly from the eigenvalue
interlacing theorem: $\lambda_1 \leq \eigmink{\COVM}{k} \leq \lambda_{n-k+1}$.

One case in which good lower bounds on \eigmink{\COVM}{k} can possibly
be obtained is when only a small (constant) number of the $\lambda_i$
are small. The following lemma allows a bound in terms of any $\lambda_j$;
however, since the running time by the implied algorithm is
exponential in $j$, and the quality of the bound depends on
$\lambda_j$, it is useful only in the special case when
$\lambda_j \gg 0$ for a small constant $j$.

\begin{lemma}
Let $V_j$ be the vector space spanned by the eigenvectors
$\Vector{e_1}, \Vector{e_2}, \ldots, \Vector{e_j}$, and define
\begin{align*}
\beta_j 
& = \max_{\Vector{y} \in V_j, \Vector{x} \in \R^n,
          \Norm[2]{\Vector{y}} = \Norm[2]{\Vector{x}} = 1,
          \Norm[0]{\Vector{x}} \leq k}
\Abs{\Vector{x} \cdot \Vector{y}}.
\end{align*}
Then, $\eigmink{\COVM}{k} \geq \lambda_{j+1} \cdot (1 - \beta_j)$.
\end{lemma}

\begin{proof}
Let $\Vector{x'} \in \R^n, \Norm[2]{\Vector{x'}} = 1, \Norm[0]{\Vector{x}} \leq k$
be an eigenvector corresponding to $\eigmink{\COVM}{k}$.
Let $\alpha_i$ be the coefficients of the representation of
\Vector{x'} in terms of the \Vector{e_i}:
$\Vector{x'} = \sum_{i=1}^n \alpha_i \Vector{e_i}$.
Thus, $\sum_{i=1}^n \alpha_i^2 = 1$, and we can write
\[ 
\eigmink{\COVM}{k}
\; = \; \Transpose{\Vector{x'}} \COVM \Vector{x'}
\; = \; \sum_{i=1}^n \alpha_i^2 \lambda_i 
\; \geq \; \lambda_{j+1} (1 - \sum_{i=1}^j \alpha_i^2).
\]
Since $\sum_{i=1}^j \alpha_i^2$ is the length of the projection of
\Vector{x} onto $V_j$, we have
\[ 
\sum_{i=1}^j \alpha_i^2
\; = \; \max_{\Vector{y} \in V_j, \Norm[2]{\Vector{y}}=1} \Abs{\Vector{x'} \cdot \Vector{y}}
\; \leq \;
\max_{\Vector{y} \in V_j, \Vector{x} \in \R^n,
\Norm[2]{\Vector{x}} = \Norm[2]{\Vector{y}} = 1, \Norm[0]{\Vector{x}} \leq k}
\Abs{\Vector{y} \cdot \Vector{x}},
\]
completing the proof.
\end{proof}

Since all the $\lambda_j$ can be computed easily, the crux in using
this bound is finding a good bound on $\beta_j$. Next, we show a PTAS
(Polynomial-Time Approximation Scheme) for approximating $\beta_j$, for
any constant $j$.

\begin{lemma}
For every $\epsilon > 0$, there is a $1-\epsilon$ approximation
for calculating $\beta_j$, running in time $O((\frac{n}{\epsilon})^j)$.
\end{lemma}

\begin{proof}
Any vector $\Vector{y} \in V_j$ with $\Norm[2]{\Vector{y}}=1$
can be written as $\Vector{y} = \sum_{i=1}^j \eta_i \Vector{e_i}$
with $\eta_i \in [-1,1]$ for all $i$.
The idea of our algorithm is to exhaustively search over all
$\Vector{y}$, as parametrized by their $\eta_i$ entries.
To make the search finite, the entries are discretized to multiples of
$\delta = \epsilon \cdot \sqrt{k/(nj)}$.
The total number of such vectors to search over is
$(2/\delta)^j \leq (n/\epsilon)^j$.

Let $\Vector{\hat{x}}, \Vector{\hat{y}}$ attain the maximum in the
definition of $\beta_j$, and write
$\Vector{\hat{y}} = \sum_{i=1}^j \hat{\eta}_i \Vector{e_i}$.
For each $i$, let $\eta_i$ be $\hat{\eta}_i$, rounded to
the nearest multiple of $\delta$, and
$\Vector{y}  = \sum_{i=1}^j \eta_i \Vector{e_i}$.
Then, $\Norm[2]{\Vector{\hat{y}} - \Vector{y}} \leq
\Norm[2]{\delta \sum_{i=1}^j \Vector{e_j}} = \delta \sqrt{j}$.

The vector $\Vector{x'} = \argmax_{%
\Vector{x} \in \R^n, \Norm[2]{\Vector{x}}=1, \Norm[0]{\Vector{x}} \leq k}
\Abs{\Vector{y}\cdot \Vector{x}}$ is of the following form:
Let $I$ be the set of $k$ indices $i$ such that $\Abs{y_i}$ is
largest, and $\gamma = \sqrt{\sum_{i \in I} y_i^2}$.
Then, $x'_i = 0$ for $i \notin I$ and $x'_i = y_i/\gamma$ for $i \in I$.
Notice that given $\Vector{y}$, we can easily find $\Vector{x'}$, and
because $\Abs{\Vector{\hat{x}}\cdot \Vector{y}}
\leq \Abs{\Vector{x'}\cdot \Vector{y}}
\leq \Abs{\Vector{\hat{x}}\cdot \Vector{\hat{y}}}$, we have
\[ 
\frac{\Abs{\Abs{\Vector{\hat{x}}\cdot \Vector{\hat{y}}} -
\Abs{\Vector{x'}\cdot \Vector{y}}}}{%
\Abs{\Vector{\hat{x}}\cdot \Vector{\hat{y}}}}
\; \leq \;
\frac{\Abs{\Abs{\Vector{\hat{x}}\cdot \Vector{\hat{y}}} -
\Abs{\Vector{\hat{x}}\cdot \Vector{y}}}}
{\Abs{\Vector{\hat{x}}\cdot \Vector{\hat{y}}}}
\; \leq \;
\frac{\Norm[2]{\Vector{\hat{x}}} \Norm[2]{\Vector{\hat{y}} - \Vector{y}}}
{\Abs{\Vector{\hat{x}}\cdot \Vector{\hat{y}}}}
\; \leq \; \frac{\delta \sqrt{j}}{\Abs{\Vector{\hat{x}}\cdot \Vector{\hat{y}}}} 
\;  \leq \; \delta \sqrt{jn/k}.
\]
The last inequality follows since the sum of the
$k$ largest entries of $\Vector{\hat{y}}$ is at least $k/\sqrt{n}$, so
by setting $x_i = 1/\sqrt{k}$ for each of those coordinates, we can
attain at least an inner product of $\sqrt{k/n}$, and the inner
product with $\Vector{\hat{x}}$ cannot be smaller.

The value output by the exhaustive search over all discretized values
is at least $\Abs{\Vector{x'} \cdot \Vector{y}}$, and thus within
a factor of $1-\frac{ \delta \sqrt{j} n}{k} = 1- \epsilon$ of the
maximum value, attained by $\Vector{\hat{x}}, \Vector{\hat{y}}$.
\end{proof}

\end{appendix}

\end{document}